\documentclass[journal]{IEEEtran}
\ifCLASSINFOpdf
\else
\fi
\usepackage{amsmath,amssymb, amsfonts}
\usepackage[pdftex]{graphicx}
\usepackage{subfig}
\usepackage{color}
\usepackage{multirow}
\usepackage[colorlinks]{hyperref}

\begin{document}

\title{BDANet: Multiscale Convolutional Neural Network with Cross-directional Attention for Building \\ Damage Assessment from Satellite Images}
%
%
%

\author{Yu Shen, Sijie Zhu, Taojiannan Yang, Chen Chen,~\IEEEmembership{Member,~IEEE,} Delu Pan, \\ Jianyu Chen,~\IEEEmembership{Member,~IEEE,}  Liang Xiao, ~\IEEEmembership{Member,~IEEE,} Qian Du, ~\IEEEmembership{Fellow,~IEEE}
\thanks{ }
}

%
%

\markboth{Journal of \LaTeX\ Class Files}%
{Shell \MakeLowercase{\textit{et al.}}: Bare Demo of IEEEtran.cls for IEEE Journals}

\maketitle

\begin{abstract}
Fast and effective responses are required when a natural disaster (e.g., earthquake, hurricane, etc.) strikes. Building damage assessment from satellite imagery is critical before relief effort is deployed. With a pair of pre- and post-disaster satellite images, building damage assessment aims at predicting the extent of damage to buildings. With the powerful ability of feature representation, deep neural networks have been successfully applied to building damage assessment. Most existing works simply concatenate pre- and post-disaster images as input of a deep neural network without considering their correlations. In this paper, we propose a novel two-stage convolutional neural network for Building Damage Assessment, called BDANet. In the first stage, a U-Net is used to extract the locations of buildings. Then the network weights from the first stage are shared in the second stage for building damage assessment. In the second stage, a two-branch multi-scale U-Net is employed as backbone, where pre- and post-disaster images are fed into the network separately. A cross-directional attention module is proposed to explore the correlations between pre- and post-disaster images. Moreover, CutMix data augmentation is exploited to tackle the challenge of difficult classes. The proposed method achieves state-of-the-art performance on a large-scale dataset -- xBD. The code is available at https://github.com/ShaneShen/BDANet-Building-Damage-Assessment. 
\end{abstract}


\IEEEpeerreviewmaketitle

\section{Introduction}
\label{SecIntro}
\IEEEPARstart{N}{atural} disasters, such as earthquakes, floods and tsunami, can cause serious social and economic devastation. When a natural disaster strikes, accurate and immediate responses are required in Humanitarian Assistance and Disaster Response (HADR) for saving thousands of lives \cite{pierdiccaTripleCollocationAssess2018, yamaguchiSensitiveDamageDetection2019, graff2020wildfire, engelSeasonalWindowEnsembleBasedThresholding2020}.  Before these responses, rescue planning and preparations are conducted based on damage analysis. With the rapid development of remote sensing technology, high resolution satellite images are now available for damage assessment. Traditionally, these images of disaster areas are analyzed by experts, which may be time-consuming if the areas are large. Therefore, automatic information extraction from satellite images, such as building segmentation and damage assessment, is imperative under time-critical situations. 


Building damage assessment plays a pivotal role in HADR, which aims at predicting the damage level for each pixel based on building segmentation. It can be divided into two categories according to the use of pre-disaster images. For the methods only using post-disaster images, the damage assessment is considered as a segmentation task \cite{ciAssessmentDegreeBuilding2019,nexStructuralBuildingDamage2019}. 
However, without pre-disaster images, building assessment errors may be inevitable since post-disaster images cannot provide precise contours of intact building objects.  With  pre-disaster images, well-shaped buildings can be utilized for damage assessment. Note the problem of damage assessment with paired images shares some similarities with change detection, as both tasks aim to find the changed regions according to two images acquired at different time \cite{zhuChangeDetectionUsing2017,tewkesburyCriticalSynthesisRemotely2015, chinIntelligentRealTimeEarthquake2020}. Compared with change detection, building damage assessment is a more challenging task because it is required to classify different damage levels. \textcolor{black}{Moreover, the unchanged objects (undamaged buildings) are required in building damage assessment, which is not considered in change detection}.


Recently, deep learning-based methods, such as convolutional neural networks (CNN), have shown their effectiveness in various tasks. With a set of labeled data, deep learning-based methods can automatically learn image feature representations from low level to high level, without selecting hand-crafted image features. Many researchers have leveraged deep neural networks for building damage assessment and achieved significant progress  \cite{jiFullyConvolutionalNetworks2019,xu2019building,weber2020building}. 
Xu et al. \cite{xu2019building} investigated the capability of CNN for building damage detection by identifying damaged and undamaged buildings.
Weber et al. \cite{weber2020building} considered building damage assessment as a semantic segmentation task, in which damage levels are assigned to different class labels.

With a pair of pre- and post-disaster images for building damage assessment, a key question is how to effectively model the correlations between these images.
Standard practice is to use a two-branch CNN-like architecture with feature fusion schemes. For instance, Hao et al. \cite{hao2020attention} concatenated the features from pre- and post-disaster images and fed them into a CNN-based framework. 
Gupta et al. \cite{gupta2020rescuenet} developed a framework that uses the difference between pre- and post-disaster features as input of the deep neural network, which is denoted as RescueNet. More recently, attention mechanism has been used with deep neural networks for remote sensing images processing \cite{liuAFNetAdaptiveFusion2020, wangMultiscaleVisualAttention2019}, which is a strategy of allocating larger weights to  informative parts of an image/feature. For instance, the average pooling is applied to extract the channel attention information for remote sensing image segmentation \cite{luoHighResolutionAerialImages2019}, and a non-local-based attention module is utilized in the network to explore the spatial relations of satellite images, in \cite{hao2020attention}.

\begin{figure}[t]
	\centering
	\subfloat{
		\includegraphics[width=0.88\linewidth]{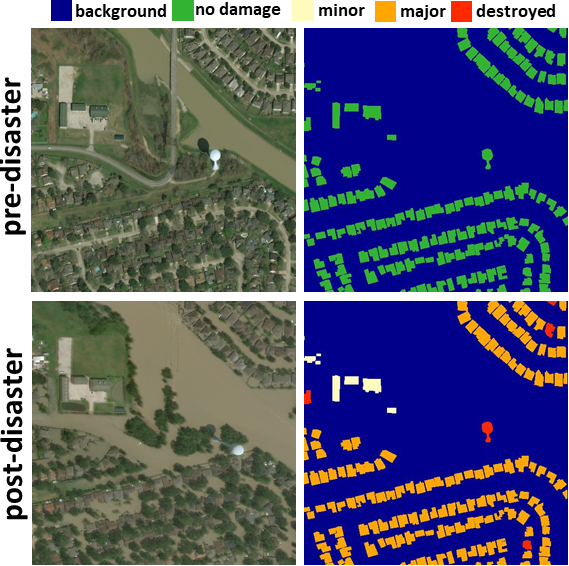}
		\label{fig:1} } 
	\caption{A pair of images and their annotations from the xBD dataset.}
	\label{figImageInstance}
\end{figure}

Another challenge of building damage assessment from satellite imagery lies in the visual similarity between certain classes (e.g., \textit{no damage} and \textit{minor damage}). These classes are considered as difficult classes. To illustrate this problem, we use the xBD \cite{gupta2019xbd} dataset, which is the largest dataset for building damage assessment to date, as an example. Fig. \ref{figImageInstance} shows a pair of images from this dataset. Based on visual observation, it is difficult to distinguish between classes such as \textit{no damage} and \textit{minor damage} due to similar appearance. Table \ref{tabBase} reports the classification results of  ResNet-50 on xBD. From the classification confusion matrix, about 24.2\% of \emph{minor damage} are mis-classified as \emph{no damage}. 
One effective strategy to cope with difficult classes and improve the model performance is data augmentation \cite{shorten2019survey,cubuk2019autoaugment}. Data augmentation has been widely used as a pre-processing technique to artificially increase the size of dataset in segmentation tasks \cite{ myronenko20183d}. Recently, CutMix \cite{yun2019cutmix} is proposed as a new data augmentation technique, which generates a new image by combining two image samples, to enhance the generalization ability of neural networks. CutMix directly cuts and pastes image patches from one image to another, which can be easily used in many tasks.

\begin{table}[t]
	\centering
	\caption{Classification confusion matrix (\%)  of ResNet-50 baseline on xBD testing set.}
	\begin{tabular}{rccccc}
	\hline
	\hline
	  Damage Level  & C0 & C1 & C2  & C3 & C4 \\
    \hline
    Background (C0) &\textbf{98.6} &	0.9 &	0.2&	0.2 &	0.1 \\
    No damage (C1) & 7.1&	\textbf{88.7} &	3.2&	0.8&	0.1\\
    Minor Damage (C2) & 6.3 &	\textcolor{blue}{24.2}&	\textbf{60.0} &	9.2&	0.4\\
    Major Damage (C3)  & 3.0 &	6.6	& \textcolor{blue}{14.9}&	\textbf{73.1} &	2.4\\
    Destroyed (C4) & 5.5 &	2.4&	1.2&	8.9&	 \textbf{82.1} \\
\hline
\hline
	\end{tabular}
\label{tabBase}
\end{table}

Motivated by the above observations, we introduce a two-stage CNN-based framework, named BDANet,  for building damage assessment. First, a single U-Net \cite{UNet} is used for building segmentation (Stage 1). U-Net is an encoder-decoder network architecture that has been widely used in segmentation tasks, e.g., in \cite{liuMultiscaleUShapedCNN2020}. 
The results of building segmentation are used to guide the locations of building damages in the second stage.
Then a two-branch multi-scale CNN (U-Net structure) is applied for damage assessment (Stage 2). By using the network weights from building segmentation, the network in Stage 2 can be trained more efficiently. Due to the scale variance of building objects, we introduce a multi-scale feature fusion (MFF) module in the encoder layer, which enables the network to aggregate features from multiple scales. In addition, a cross-directional attention (CDA) module is proposed to explore the correlations between features from pre- and post-disaster images. By leveraging channel-wise and spatial-wise correlations, the network can fuse and enhance the features from pre- and post-disaster images. Moreover, to tackle difficult classes, CutMix is employed for data augmentation. 
Specifically, we only apply CutMix to difficult classes to make the network pay more attention to those classes, thereby learning more robust representations for their separation. In the experiments, we show that this strategy yields superior classification performance over simply adopting CutMix for all classes. 

The contributions of this paper are summarized as follows.
\begin{itemize}
\item We present a two-stage CNN-based learning framework for building segmentation (Stage 1) and damage assessment (Stage 2). The building segmentation helps the network to locate the building objects. The proposed framework achieves state-of-the-art performance on a large-scale building damage assessment benchmark -- xBD \cite{gupta2019xbd}.

\item To reduce the influence of  scale variance of buildings, a multi-scale feature fusion (MFF) module is introduced to learn features from multiple scales.

\item A cross-directional attention (CDA) module is proposed to effectively aggregate the features from pre- and post-disaster images. With the attention module, informative channel and spatial information can be extracted to enhance the feature representation.

\item We also unveil the challenge of difficult classes in building damage assessment and explore an effective data augmentation strategy, CutMix \cite{yun2019cutmix}, to cope with this challenge. 

\end{itemize}

The remainder of this paper is organized as follows. Section \ref{SecRelatedWork} gives an overview of related works.  Section \ref{SecMethod} presents the details of the proposed BDANet for building damage assessment. Section \ref{SecExp} provides comprehensive experimental results and parameter analysis. Finally, conclusion is drawn in Section \ref{SecConclusion}.

\section{Related Work}
\label{SecRelatedWork}
\subsection{Building Damage Assessment}
\textcolor{black}{Building damage-based analysis, including detection, segmentation and assessment, are vital topics in HADR. Recently, many algorithms have been developed for building damage analysis \cite{dongComprehensiveReviewEarthquakeinduced2013,anniballeEarthquakeDamageMapping2018,kalantarAssessmentConvolutionalNeural2020}. For instance, in \cite{anniballeEarthquakeDamageMapping2018}, to detect the damaged building objects after an earthquake, change features, including texture, color, statistical similarity and correlation descriptors, are all analyzed before applying a support vector machine (SVM) classifier.
In \cite{Zhu_2021_WACV}, Zhu et. al. proposed an instance segmentation network, named MSNet, for building damage detection. In \cite{vetrivelDisasterDamageDetection2018}, a multiple kernel learning framework is proposed. To improve performance of building damage detection,  CNN feature and 3D cloud point feature are integrated in the framework. In \cite{tilonPostDisasterBuildingDamage2020}, an unsupervised anomaly detection method is developed to automatically detect the location of damaged buildings.}


\textcolor{black}{Different from building damage detection and segmentation, building damage assessment aims at identifying the damage level of building objects after natural disasters.} Depending on whether to use pre-disaster images, building damage assessment can be divided into two categories. One category only uses post-disaster images and the other uses paired pre- and post-disaster images for damage assessment.
Many previous works focus on using only post-disaster images. For example, in \cite{liUrbanBuildingDamage2009}, SVM is used to assess the building damage after earthquake with QuickBird satellite images. 
In \cite{taskinkayaDamageAssessment20102011}, object-based analysis is integrated with machine learning methods to preserve the boundary and detailed information of remote sensing satellite images in damaged buildings. However, without pre-disaster images, building assessment errors can be easily made because post-disaster images cannot provide precise boundaries/contours of complete building objects. Therefore, researchers resort to use a pair of satellite images (pre- and post-disaster) for building damage assessment. Cooner et al. \cite{coonerDetectionUrbanDamage2016} extracted  textural and structural features from both pre- and post-disaster images, which  improved the  performance of earthquake damage assessment. Due to the limited amount of labeled data, some works simplify the damage assessment task as binary classification, which assigns \emph{damage} and \emph{no damage} labels to building objects. For example, Xu et al. \cite{xu2019building} compared the performance of different CNN models in detecting damage buildings in the Haiti earthquake. 



Recently, researchers start to differentiate more building damage levels for better disaster analysis. Different from the binary classification, the extent of damage is considered, such as minor damage and major damage. In \cite{nexStructuralBuildingDamage2019}, the level of building damage after earthquake is evaluated by a framework integrating CNN and ordinal regression. 
In \cite{presa-reyesAssessingBuildingDamage2020}, to reduce the uncertainty in predicting building damage levels, building objects in the center of an image patch are considered while the surrounding buildings are occluded by negating the pixels. Hence, the model can focus on the buildings in the center of an image patch.

\begin{figure*}[!t]
	\centering
	\includegraphics[width=0.84\textwidth]{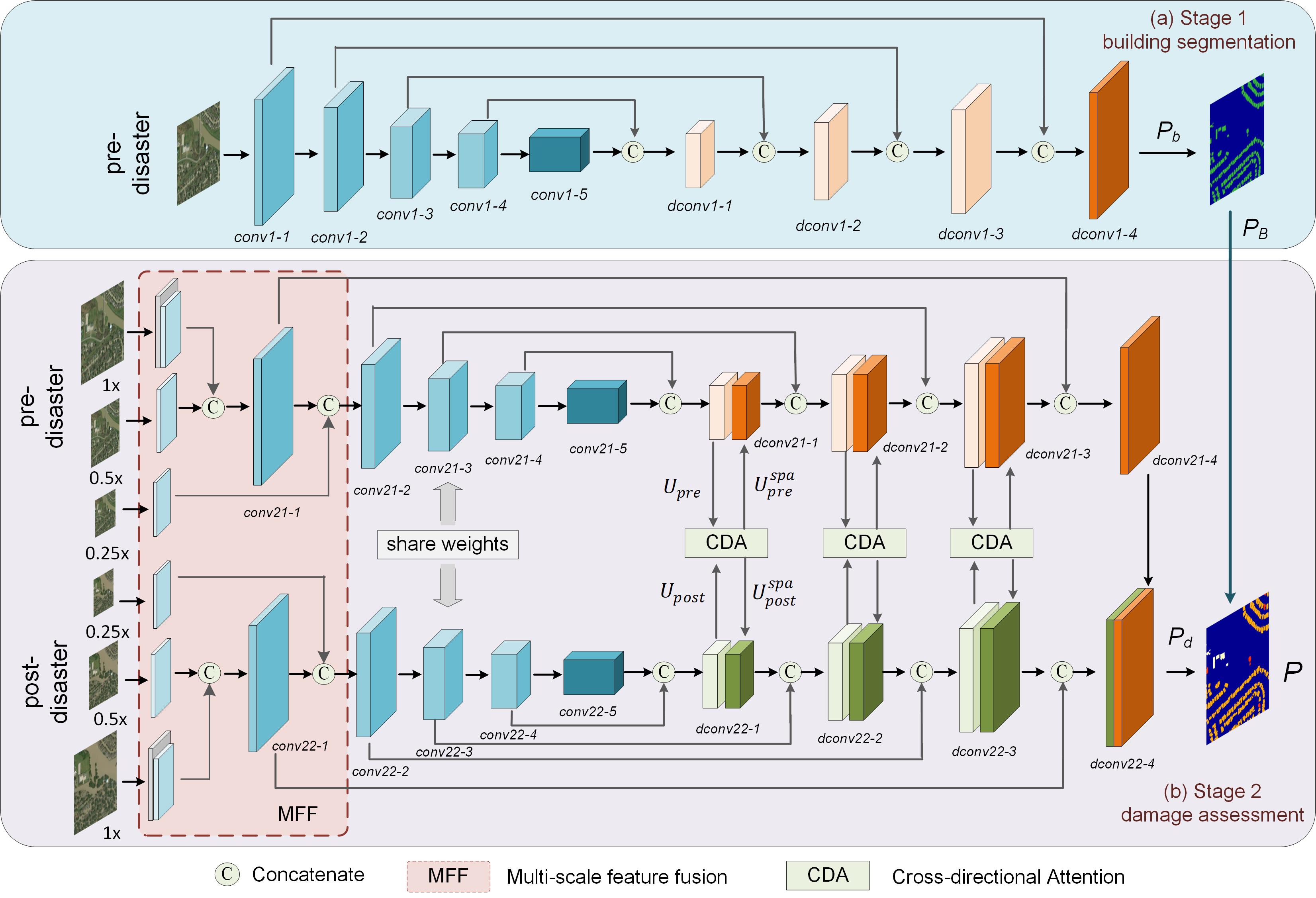}
	\caption{Overview of the proposed framework (BDANet). The U-Net structure is used in both stages. It consists of two stages: (a) Stage 1: building segmentation, (b) Stage 2: damage assessment. }
\label{figFramework}
\end{figure*}

\subsection{U-Net}
\textcolor{black}{U-Net is  a  CNN-based  network  that  was  first  proposed for   biomedical  image  segmentation  \cite{UNet}. Compared  with  traditional  CNN-based  architectures,  U-Net introduced down-sampling and up-sampling to aggregate feature from multiple  scales.  By  adopting  the  skip  connection,  features from low levels and high levels are integrated to enhance the learning ability of network  \cite{liuSurveyUshapedNetworks2020}. Owning to these advantages, U-Nets have been widely used in image segmentation \cite{dingMultiscaleFullyConvolutional2020} and classification \cite{liGeometryattentionalNetworkALS2020}. }

\textcolor{black}{U-Net-based architectures have also been applied in remote sensing image processing. For instance, in \cite{chenSymmetricalDenseShortcutDeep2018}, U-Net is introduced for very-high resolution remote sensing image segmentation. In \cite{chenMultiScaleSpatialChannelwise2020}, U-Net is integrated with multi-scale strategy for building instance extraction. In \cite{rsLiBuildingExtraction}, U-Net is used as a feature extraction method for building segmentation, which provides promising results. Moreover, U-Net-based architectures have been applied in a variety of tasks, such as road extraction \cite{xuRoadExtractionHighResolution2018} and change detection \cite{chenDASNetDualAttentive2021}. However, the U-Net architecture is simple and cannot be directly used in building damage assessment. }

\subsection{Attention-based Models}
Attention-based models \cite{Hu_2018_CVPR,wang2018non} have been widely used in computer vision tasks, which learn to put more emphasis on important information of images/features. Recently, many attention-based models have been developed and achieved great success in remote sensing image processing tasks, such as image classification and segmentation \cite{raoBidirectionalGuidedAttention2020, chenMultiScaleSpatialChannelwise2020}. 
In \cite{luoHighResolutionAerialImages2019}, a channel-attention mechanism is integrated with a fully convolutional network (FCN) for high-resolution aerial image segmentation.  
In \cite{Zheng_2020_CVPR}, a foreground-aware network is developed, which incorporates both spatial and channel attention models to improve foreground modeling, for geospatial object segmentation. In \cite{dingLANetLocalAttention2020}, an attention embedding module is designed to embed attention from high-level layers into low-level ones to enrich the semantic information for satellite image segmentation.
In \cite{zhangPositionalContextAggregation2020}, a self-attention mechanism that considers both spatial-dipartite context aggregation
information and the relative position encoding information is proposed for remote sensing scene classification. 
 
Attention-based models are barely applied in building damage assessment. In \cite{hao2020attention}, a non-local attention model \cite{wang2018non} is adopted to capture long-range spatial information of pre- and post-disaster images. However, the computational cost of the non-local mechanism is very high because the attention map is calculated based on high-resolution features.

\subsection{Data Augmentation Strategies}
Data augmentation has been widely used as a pre-processing technique to artificially increase the size of dataset in computer vision tasks \cite{takahashiDataAugmentationUsing2020, myronenko20183d, liDataAugmentationHyperspectral2019}. Data augmentation methods usually employ image transformations, such as image flip, rotation and crop, to improve the model generalization performance. Manually annotating remote sensing images is difficult and time-consuming. As a result, labeled remote sensing data are usually limited. To facilitate the training of deep neural networks, data augmentation is critical in deep learning-based remote sensing image processing.

Recently, various data augmentation techniques have emerged and shown their efficiency. 
Cutout \cite{devries2017improved} randomly masks out square regions of input to force the network pay attention to diverse image regions, thereby improving the feature representation capability of CNNs.
Some works focus on synthesizing image samples to expand the size of training data. Cubuk et al. \cite{cubuk2019autoaugment} proposed a search algorithm for image augmentation, which enables the network to find the best validation accuracy. In \cite{zhang2018mixup}, Mixup is introduced to regularize the neural network by using linear combination of pairs of samples and labels. Different from the Mixup strategy that merges two images with linear summation, CutMix \cite{yun2019cutmix} directly cuts and pastes image patches from one image to another, which can be used in segmentation tasks.

\textcolor{black}{At present, most deep learning based methods still follow the architectures of change detection and seldom consider the completeness of building objects. Furthermore, due to the similarities between certain classes, it is difficult for the network to distinguish their differences. To deal with this problems, a two-stage framework is proposed for building damage assessment. The MMF and CDA models are proposed to enhance the feature representation. Moreover, the CutMix strategy is employed to alleviate the difficulty of classifying hard classes.}

\begin{figure*}[!t]
	\centering
	\includegraphics[width=0.64\textwidth]{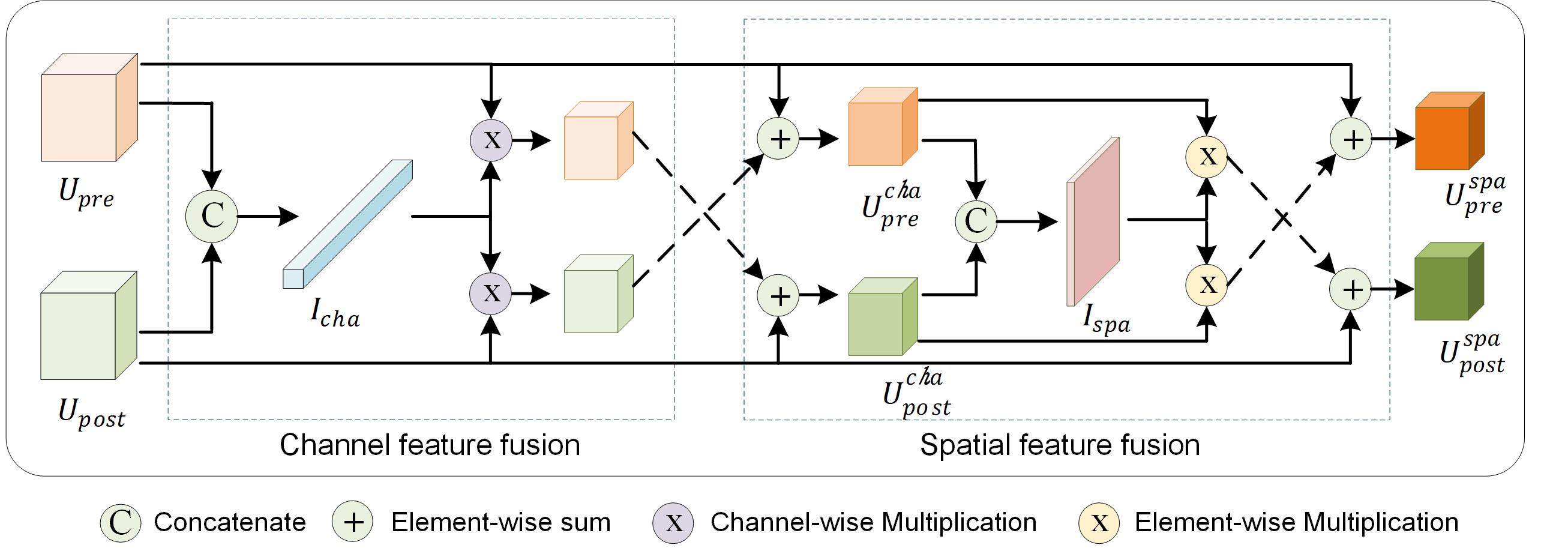}
	\caption{Framework of the proposed cross-directional attention (CDA) module. }
\label{figCross}
\end{figure*}

\section{Proposed Method}
\label{SecMethod}
Fig. \ref{figFramework} presents an overview of the proposed BDANet, which consists of two stages: building segmentation (Stage 1) and damage assessment (Stage 2). In Stage 1, a single U-Net branch (i.e., the upper one) is used for building segmentation. This U-Net branch uses only pre-disaster images as input and produces segmentation masks of building objects. In Stage 2, the pre- and post-disaster images are fed into two network branches separately. To further enhance the feature representations, the multi-scale feature fusion (MFF) module and the cross-directional attention (CDA) module are introduced. We describe the details of the proposed damage assessment framework and its components in the following. 

\subsection{Two-stage Framework} 
\textcolor{black}{Although the building damage assessment task shares some similarities with change detection task, there are several key differences. First, building damage assessment not only requires to detect the changes but also needs to classify their damage levels. Second, the undamaged buildings also need to be detected, which are usually unnecessary in change detection. Change detection-based networks mainly follow a three-branch architecture, which contains a dual-branch for image pairs learning and a fusion branch for change detection \cite{houWNetCDGANBitemporal2020, daudtFullyConvolutionalSiamese2018}. These networks are designed for learning the differences of image pairs while unchanged objects (e.g. undamaged buildings) are ignored. Due to these problems, directly using networks for change detection cannot satisfy the needs of building damage assessment. Therefore, in this paper, a two-stage learning framework is proposed for building damage assessment.}

\textbf{Stage 1: building segmentation} As shown in Fig. \ref{figFramework}(a), to locate the building objects, only pre-disaster images are used as input to the network in this stage. The network is separately trained for building objects segmentation in this stage. Compared with post-disaster images, building objects in pre-disaster images are intact and well-shaped. The network is designed based on the architecture of U-Net \cite{UNet}. In the encoder part, a ResNet \cite{he2016deep} is used as backbone to extract deep features. 
By using the strided convolution, the input image is first down-sampled into half of the original resolution. With the network becoming deeper, feature maps are projected into a higher dimension for high-level feature extraction. In the decoder part, the resolution of features are gradually recovered to the original input size by upsampling. To capture the details of target objects, skip connections are utilized in the network (see Fig. \ref{figFramework}(a)). Before each upsampling operation, the low-level feature maps from the encoder layer are concatenated with the high-level feature maps. The output of the network is a binary segmentation map indicating the building distribution in a satellite image. \textcolor{black}{Similar to some segmentation tasks, the cross-entropy loss is used in this stage, which is formulated as  
\begin{equation}
	L = -\sum_i^N{y^i\log\hat{y}^i + (1-y^i)\log(1-\hat{y}^i)},
	\label{eqCEL}
\end{equation}
where $L$ quantifies the overall loss by comparing the target label $y^i$ with predicted label $\hat{y}^i$, $N$ represents the number of training pixels.}

\textcolor{black}{The objective of Stage 1 is locating the building objects. 
As many buildings are damaged in post-disaster images, it is difficult for the network in Stage 2 to perform building segmentation and damage assessment at the same time. 
 Therefore, to guide the assessment process of Stage 2, a single U-Net using only pre-disaster images is first applied to identify the locations of  buildings. }

\textbf{Stage 2: damage assessment} In Stage 2, both pre- and post-disaster images are fed into a two-branch network.
The backbone network for damage assessment is the same as that for building segmentation. The convolutional kernels are shared in both branches. Therefore, the parameter weights obtained from Stage 1 are used here for weights initialization.  

As shown in Fig. \ref{figFramework}(b), the feature maps from two branches are concatenated in dconv4 layer. Then the output of damage assessment is generated after a convolutional layer. The cross-enropy loss is also applied in Stage 2.
Let the output of building segmentation be $P_b\in \mathbb{R}^{2\times H\times W}$, where $H$ and $W$ denote the height and width of the output, respectively.
Then the segmentation results can be expressed as
\begin{equation}
P_B = \arg\max (P_b),
\label{eqPB}
\end{equation}
where $P_B \in \{0,1\}^{1\times H\times W}$ with 1 representing for buildings and 0 for background.

Let the output of damage assessment be $P_d \in \mathbb{R}^{C\times H\times W}$, where $C$ denotes the number of damage levels. To guide the building locations, the final output $P$ is represented as
\begin{equation}
P = \arg\max (P_B \cdot P_d),
\label{eqOutput}
\end{equation}
where $\cdot$ denotes the element-wise multiplication. 

\subsection{Multi-scale Feature Fusion Module}
As the scale of buildings varies in images captured from various disasters and regions, learning a robust representation of building objects under various scales is important. To this end, a multi-scale feature fusion (MFF) module is introduced in the encoder of the network. The input image is processed with three different scales:  $1\times$ input resolution (original size), $0.5\times$ input resolution, $0.25\times$ input resolution. Features from different scales are extracted by a convolutional block and then embedded in the encoder of the network, as shown in Fig. \ref{figFramework}(b). Specifically, the MFF contains three streams. The first stream uses images with the original resolution. In this stream, a convolutional layer with stride 2 is used to generate feature maps with half of the input size. This stream is the same as the first layer of ResNet, which is used as the backbone of the proposed framework.
The $0.5\times$  resolution images are used as input of the second stream. The size of feature maps of this stream keeps the same with $0.5\times$ resolution images. Then feature maps from $1\times$ and $0.5\times$ streams are concatenated for further processing. To keep a similar setting of the number of channels as the ResNet, a dimension reduction is conducted on the concatenated feature map. In the third stream, the feature map of $0.25\times$ resolution is concatenated with the second convolutional block of ResNet. In this way, features from three different image scales are embedded in the network, which enables the network to learn a robust features.

\begin{table*}[!t]
	\centering
	\caption{Parameters of the BDANet.}
	\setlength{\tabcolsep}{2.5mm} {
	\begin{tabular}{ccc|ccccc}
		\hline \hline
		& Stage 1 & & & & Stage 2 & & \\
		\hline
		Layer  &  Feature Size &  Kernel Size & Layer  &  Feature Size  & Layer & Feature Size & Kernel Size\\
		\hline
		Input-pre &  $512\times512\times3$ & - & Input-pre & $512\times 512 \times 3$  & Input-post & $512\times 512 \times 3$  & - \\
		conv1-1 & $256\times256\times64$ & $5\times5$ & conv21-1 &$256\times256\times64$ & conv22-1  & $256\times256\times64$ & $5\times5$\\
		conv1-2 &  $128\times128\times256$ & $3\times3$  & conv21-2 & $128\times128\times256$ & conv22-2 & $128\times128\times256$ & $3\times3$\\
		conv1-3 &  $64\times64\times512$ & $3\times3$  & conv21-3 & $64\times64\times512$ & conv22-3 & $64\times64\times512$  & $3\times3$\\
		conv1-4 &  $32\times32\times1024$ & $3\times3$  & conv21-4 &  $32\times32\times1024$ & conv22-4 &  $32\times32\times1024$  & $3\times3$\\
		conv1-5 &  $16\times16\times2048$ & $3\times3$  & conv21-4 &  $16\times16\times2048$ & conv22-4 &  $16\times16\times2048$  & $3\times3$\\
		dconv1-1 &  $32\times32\times512$ & $3\times3$  & dconv21-1 &  $32\times32\times1024$ & dconv22-1 &  $32\times32\times1024$  & $3\times3$\\
		dconv1-2 &  $64\times64\times256$ & $3\times3$  & dconv21-2 &  $64\times64\times512$ & dconv22-2 &  $64\times64\times512$  & $3\times3$\\
		dconv1-3 &  $128\times128\times96$ & $3\times3$  & dconv21-3 &  $128\times128\times192$  & dconv22-3 &  $128\times128\times192$  & $3\times3$\\
		dconv1-4 &  $256\times256\times32$ & $3\times3$  & dconv21-4 &  $256\times256\times32$  & dconv22-4 &  $256\times256\times64$  & $3\times3$\\
		Output &  $512\times512\times1$ & - & - & - & Output & $512\times 512 \times 5$  & - \\
\hline	\hline	
	\end{tabular}}
	\label{tabParameters}
\end{table*}

\subsection{Cross-directional Attention Module} 
To further explore the correlations between pre- and post-disaster features, a cross-directional attention (CDA) module is developed. Inspired by the squeeze and excitation (SE) block \cite{royRecalibratingFullyConvolutional2019}, the proposed CDA module focuses on recalibrating features from channel and spatial dimensions. 
Moreover, the channel and spatial information from  pre- and post-disaster features is aggregated together in a cross manner and then embedded in the network respectively. The details are depicted in Fig. \ref{figCross}. Let $U_{pre} \in\mathbb{R}^{E\times h\times w}$ and $U_{post}\in\mathbb{R}^{E\times h\times w}$ be the feature maps obtained from the two branches of U-Net respectively, where $E$ denotes the number of channels of feature maps, $h$ and $w$ are the height and width of feature maps, respectively. Then the channel information can be extracted by
\begin{equation}
I_{cha} = \sigma(P_g([U_{pre}, U_{post}])), 
\label{eqAverageP}
\end{equation}
where $[U_{pre}, U_{post}]$ denotes the concatenation of feature maps, $P_g(\cdot)$ represents the global average pooling,  $\sigma(\cdot)$ is the sigmoid function, and $I_{cha}$
is a feature vector of $E$ channels after dimension reduction from $2E$. Then the new features from two branches can be formulated as
\begin{equation}
\begin{aligned}
U_{pre}^{cha} = I_{cha}\ast U_{post} + U_{pre}, \\
U_{post}^{cha} = I_{cha}\ast U_{pre} + U_{post},
\end{aligned}
\label{eqChannel}
\end{equation}
where $\ast$ denotes the channel-wise multiplication between the input feature maps and vector $I_{cha}$. The output of channel feature fusion is then used in the spatial feature fusion. We concatenate $U_{pre}^{cha}$ and $U_{post}^{cha}$, and feed them into a $1\times1$ convolution $P_{conv}$ as
\begin{equation}
I_{spa} = \sigma(P_{conv}([U_{pre}^{cha}, U_{post}^{cha}])), 
\label{eqConv1}
\end{equation}
where $P_{conv}([U_{pre}^{cha}, U_{post}^{cha}]) \in \mathbb{R}^{1\times H \times W}$.
Then the output features are 
\begin{equation}
\begin{aligned}
U_{pre}^{spa} = I_{spa}\cdot U_{post}^{cha} + U_{pre}, \\
U_{post}^{spa} = I_{spa}\cdot U_{pre}^{cha} + U_{post},
\end{aligned}
\end{equation}
where $I_{spa}\cdot U_{post}^{cha}$ and $I_{spa}\cdot U_{pre}^{cha}$ denote the spatial element-wise multiplication. As a result, channel and spatial information from pre- and post-disaster branches are effectively aggregated in the CDA module.
It is  worth noting that the proposed CDA module consists of only simple convolution and matrix operations, which is easy to implement and integrate with existing CNN architectures. 

\textcolor{black}{Therefore, the overall architecture of BDANet is established. The detailed parameters are listed in Table \ref{tabParameters}.}

\begin{figure}[h]
	\centering
	\subfloat{
		\includegraphics[width=0.99\linewidth]{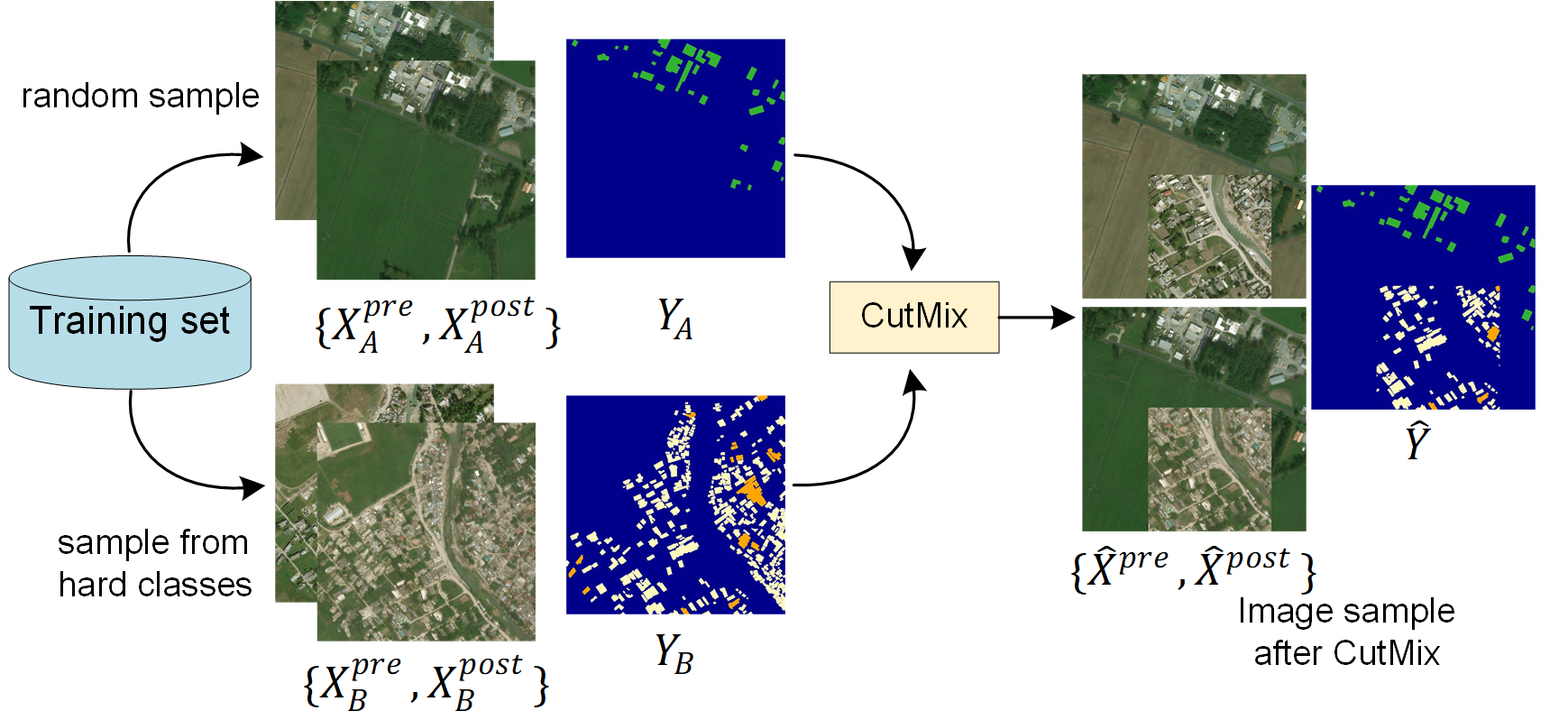}
	\label{fig:1} } 
	\caption{Data augmentation with CutMix for difficult classes.}
	\label{figCutmix}
\end{figure}

\subsection{Data Augmentation with CutMix} 
As discussed in Section \ref{SecIntro}, several damage levels are difficult to distinguish from each other due to their similarities in satellite images. To address this challenge, we leverage the CutMix \cite{yun2019cutmix} data augmentation scheme to increase the sample sizes of difficult classes, for better feature representations.  
Specifically, image patches are cut from samples that contain difficult classes (i.e., \textit{minor} and \textit{major} damage classes for the xBD dataset based on the results in Table \ref{tabBase}), and then pasted into any random sample images. The CutMix procedure is illustrated in Fig. \ref{figCutmix}. Let $\{X_A^{pre}, X_A^{post}\}  \in \mathbb{R}^{c\times H\times W}$ and $Y_A \in \mathbb{R}^{H\times W}$ be a randomly selected training sample (an image pair) and label, ($X_B^{pre}, X_B^{post}, Y_B$) be a randomly selected sample from difficult classes, where $c$ is the channel number of images. Then the CutMix operation in this task can be defined as
\begin{equation}
\begin{aligned}
\hat{X}^{pre} = M\cdot X_A^{pre} + (1-M) \cdot X_B^{pre} , \\
\hat{X}^{post} = M\cdot X_A^{post} + (1-M) \cdot X_B^{post}, \\
\quad \hat{Y} = M\cdot Y_A + (1-M) \cdot Y_B,  
\end{aligned}
\label{eqConv1}
\end{equation}
where $M\in \mathbb{R}^{1\times W\times H} $ denotes a binary mask indicating where to cut out and fill in from two image samples,  and $(\hat{X}^{pre}, \hat{X}^{post}, \hat{Y})$ represents the generated new sample. With the increased sample size using CutMix, the network is forced to pay more attention to these difficult classes and learn more robust representations.


\section{Experiments}
\label{SecExp}
\subsection{Dataset and Evaluation}
To evaluate the effectiveness of our proposed method, the xBD dataset \cite{gupta2019xbd} is used in the experiments. The xBD dataset is the only public available large-scale dataset of satellite images for building segmentation and damage assessment at present. It was sourced from the DigitalGlobe Open Program\footnote{https://www.digitalglobe.com/ecosystem/open-data}, which was released for sudden major disaster events. Specifically, the Open Data Program was chosen for its availability of high-resolution imagery from many disparate regions of the world. In this dataset, 19 different disasters (such as hurricanes, floods, wildfire and earthquakes) are selected at various locations with more than 800,000 building annotations. The dataset consists of image pairs (pre- and post-disaster) of size $1024 \times 1024$ pixels. Each image contains three spectral bands (red, green, blue) and has a resolution of 0.8 meter per pixel. The damage assessment contains 4 levels, including no damage, minor damage, major damage and destroyed. The training and testing sets are listed in Table \ref{tabDatasplit}. It is worth mentioning that the data is imbalanced and the damage level is highly skewed toward \textit{no damage}. The number of each damage level's polygons is reported in Table \ref{tabLevelDis}.

The experimental results are evaluated by using the F1 score ($F1_b$) for building segmentation  and the harmonic mean of class-wise damage classification F1 ($F1_d$) for building damage assessment, which are defined as:
\begin{equation}
\begin{aligned}
F1_b = \frac{2TP}{2TP+FP+FN}, \\
F1_d = \frac{n}{\sum_{i=1}^{n}1/F1_{C_i}}, \\
\end{aligned}
\label{eqF1}
\end{equation}
respectively, where $TP$, $FP$ and $FN$ are the number of true-positive, false-positive and false-negative pixels of segmentation results, respectively, $ 1/F1_{C_i}$ denotes the F1 score of each damage level for damage assessment, and $C_i$ denotes the damage level. F1 score is a challenging evaluation index as it heavily penalizes the overfitting to over-represented classes. The overall score $F1_s$ \cite{gupta2019xbd} comprehensively evaluates the performance of building segmentation and damage assessment, which
is formulated by a weighted average of $F1_b$ and $F1_d$:
\begin{equation}
F1_s = 0.3 \times F1_b + 0.7 \times F1_d.
\label{eqOveral}
\end{equation}

\begin{table}[tp]
	\centering
	\caption{The xBD dataset splits and annotation numbers. }
	\setlength{\tabcolsep}{2.6mm} {
		\begin{tabular}{c|cccccccc}
			\hline \hline
    {Split}     & {Image No.}     & {Polygon No.} \\
    \hline
    Train & 18336  &  632228     \\
    Test    & 1866  & 109724     \\
\hline	\hline	
	\end{tabular}}
	\label{tabDatasplit}
\end{table}

\begin{table}[tp]
	\centering
	\caption{Damage level annotations and the distribution.}
	\setlength{\tabcolsep}{2.6mm} {
		\begin{tabular}{c|cccc}
			\hline \hline
       & No damage    & Minor & Major & Destroyed \\
    \hline
   \textbf{No.} & 313003  &  36860    & 29904 & 31560 \\
   \%  &     76.04     & 8.98 & 7.29 &7.69              \\
\hline	\hline	
	\end{tabular}}
	\label{tabLevelDis}
\end{table}

\begin{table*}[!t]
	\centering
	\caption{Building damage assessment results on xBD dataset with different methods. }
	\setlength{\tabcolsep}{2.6mm} {
		\begin{tabular}{r|cccccccc}
			\hline \hline
	Method & $F1_s$ (overall) & $F1_b$ & $F1_d$  & No damage    & Minor & Major & Destroyed \\
	\hline
	WNet \cite{houWNetCDGANBitemporal2020} & 0.737&	0.817&	0.703	&0.884	&0.518&	0.684&	0.855 \\
	U-Net++ \cite{pengEndtoEndChangeDetection2019} & 0.740&	0.819	&0.707&	0.886&	0.523&	0.689&	0.858\\
       \hline
    RescueNet \cite{gupta2020rescuenet} (2020) & 0.741 & 0.835 &	0.697 &	0.906 &	0.493 &	0.722 &	0.837 \\
   Weber et al. \cite{weber2020building} (2020)  & 0.770 &	0.840 &	0.740 &	0.885 &	0.563 &	0.771 &	0.808   \\
   \hline
   FCN \cite{Long_2015_CVPR} & 0.765&	0.864&	0.722&	0.919&	0.532&	0.708&	0.861 \\
   SegNet \cite{badrinarayananSegNetDeepConvolutional2017} & 0.782&	0.864&	0.747&	0.921&	0.567&	0.746&	0.859
\\
   DeepLabv3 \cite{chen2017rethinking} &0.788&	0.864&	0.756&	0.919&	0.580&	0.761&	0.859
 \\
   Ours (vanilla network) & 0.789 & 0.864 &	0.757 &	0.923 &	0.578 &	0.760 &	0.869  \\
   Ours (BDANet) &  \textbf{0.806} &	\textbf{0.864} & \textbf{0.782} &	\textbf{0.925} &	\textbf{0.616} &	\textbf{0.788} &	\textbf{0.876}   \\
\hline	\hline	
	\end{tabular}}
	\label{tabResults}
\end{table*}

\begin{table}[t]
	\centering
	\caption{Classification confusion matrix (\%)  of proposed BDANet on xBD testing set.}
	\begin{tabular}{rccccc}
	\hline
	\hline
	  Damage Level  & C0 & C1 & C2  & C3 & C4 \\
    \hline
    Background (C0) &\textbf{98.7} 	&0.7 &0.3 &	0.2 &	0.1  \\
    No damage (C1) & 6.7 	&\textbf{88.8} &3.4 & 1.0 &	0.2 \\
    Minor Damage (C2) & 5.3 &18.5 &\textbf{64.5} &11.2 &0.5\\
    Major Damage (C3)&  2.9 &	4.8 &	12.6 &	\textbf{77.8} &	1.9\\
    Destroyed (C4) &5.9 &	1.8 &	0.7 &	8.9 &	\textbf{82.7} \\
\hline
\hline
	\end{tabular}
\label{tabConfusion}
\end{table}

\subsection{Implementation Details}
We implement the proposed framework using Pytorch. Experiments are conducted on a computer with an Intel i9-9920X CPU and two NVIDIA TITAN-V GPUs. Images are cropped to a size of $512\times 512$ pixels for training. The prediction for the testing set is performed on the original size of images ($1024\times 1024$). Apart from CutMix, basic data augmentation is conducted in the training phase, such as flip and rotation. Due to GPU memory limitation, we use ResNet-50 as the backbone for all compared networks with pre-trained weights loaded from the PyTorch library. The number of channels in each convolutional block is [64, 256, 512, 1024, 2048]. In the decoder, the number of channel in each convolutional block is [512, 256, 96, 32]. The number of convolutional channels are the same as  Stage 1 and Stage 2. The cross-entropy loss is used for both building segmentation and damage assessment. The optimization method is Adam. In the building segmentation stage, the learning rate is 0.00015 and the number of epoch is 120. In the damage assessment stage, the learning rate is 0.0002 and the number of epoch is 25.

\subsection{Results Analysis}
To validate the effectiveness of the proposed framework, we employ several state-of-the-art methods for comparison on xBD. 

\textcolor{black}{
1) Two change detection-based networks, including WNet \cite{houWNetCDGANBitemporal2020} and U-Net++ \cite{pengEndtoEndChangeDetection2019} are evaluated for building damage assessment.}

2) The RescueNet \cite{gupta2020rescuenet} uses a dilated ResNet-50 as the backbone network. The difference of features from pre- and post-disaster images is utilized for damage assessment.

3) In Weber's method \cite{weber2020building}, after feature extraction with ResNet-50, features from pre- and post-disaster images are concatenated for feature fusion. Then the segmentation is performed for damage assessment.

4) Several state-of-the-art semantic segmentation networks, including FCN \cite{Long_2015_CVPR}, SegNet \cite{badrinarayananSegNetDeepConvolutional2017} and DeepLabv3 \cite{chen2017rethinking}, are also employed for comparison. We set the two-stage framework unchanged and only replace the U-Net-based architecture in Stage 2 with these mentioned networks. According to Table \ref{tabResults}, both methods provided low scores  compared with other networks. It is reasonable 

Apart from our proposed two-stage framework, a network without the multi-scale feature fusion (MFF) module, cross-directional attention (CDA) module and CutMix is used as a clean baseline for comparison, which is denoted as the \textbf{vanilla network}. The damage assessment results of different approaches are reported in Table \ref{tabResults}. The proposed framework using the vanilla network alone outperforms the existing methods, e.g., \cite{gupta2020rescuenet} and \cite{weber2020building}, in terms of all three metrics ($F1_s$, $F1_b$ and $F1_d$). Unlike the RescueNet and Weber's method, the decoder and skip connections are used in  our framework. Therefore, our method provides a better performance in the building segmentation stage (i.e., an improvement of over 2\% in $F1_b$). 
Moreover, our method achieves a significant improvement (about 12\%, 0.493$\rightarrow$0.616) in \textit{minor damage} level compared with RescueNet \cite{gupta2020rescuenet}.

\textcolor{black}{Different from our proposed framework, the change detection-based networks (WNet \cite{houWNetCDGANBitemporal2020} and U-Net++ \cite{pengEndtoEndChangeDetection2019}) perform buidling segmentation and damage assessment in one network. Accodring to Table \ref{tabResults}, low scores of $F1_s$ are provided by both methods. It is reasonable because is difficult for networks to locate buildings when post-damage images are also fed into the networks. From Table \ref{tabResults}, WNet provided the lowest score on $F1_b$ and the \textit{no damage} level.}
As FCN uses simple upsampling to predict the damage levels and hardly considers features from shadow layers, the score in $F1_s$ is not satisfactory compared with other methods under the same framework. In contrast, SegNet and DeepLabv3 have achieved promising results using our proposed framework. Compared with the vanilla network, our overall approach incorporating the proposed components further improves the accuracy for damage assessment (0.782 vs 0.757 in $F1_d$), and the improvement is consistent for all the damage levels. By considering both the building segmentation and damage assessment performance (i.e., the overall score $F1_s$), our proposed two-stage framework outperforms the existing methods \cite{gupta2020rescuenet,weber2020building} by a considerable margin. 
\textcolor{black}{The classification confusion matrix of proposed framework is reported in Table \ref{tabConfusion}. Compared with classification accuracy in Table \ref{tabBase}, much improvements are achieved in the \textit{minor damage} and \textit{major damage} level, indicating the effectiveness of our proposed method. }

\begin{figure}[t]
	\centering
	\subfloat{
		\includegraphics[width=0.955\linewidth]{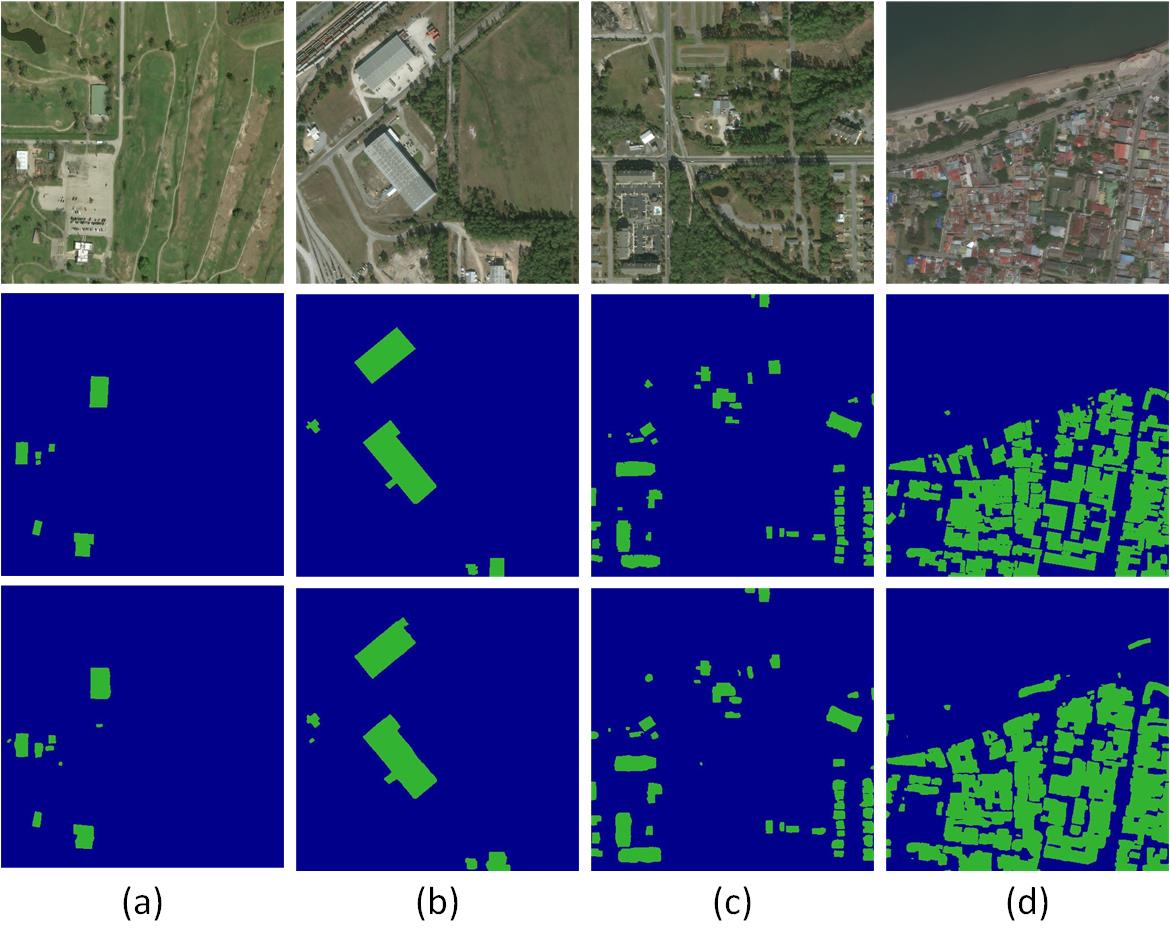}
		\label{fig:1} } \hspace{-3mm}
	\caption{Visual examples of building segmentation in Stage 1. First row to third row are: pre-disaster images, ground truth of building segmentation, and segmentation results of Stage 1. }
	\label{figBuildingloc}
\end{figure}

\begin{figure*}[t]
	\centering
	\subfloat{
		\includegraphics[width=0.130\linewidth]{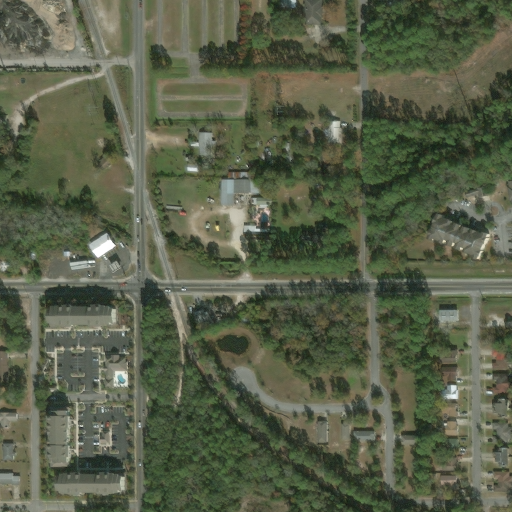}
		\label{fig:1} } \hspace{-2.5mm}  
	\subfloat{
		\includegraphics[width=0.130\linewidth]{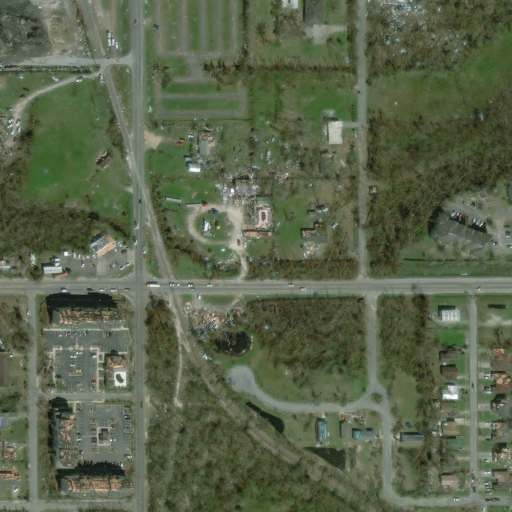}
		\label{fig:2}}    \hspace{-2.5mm}  
	\subfloat{
		\includegraphics[width=0.130\linewidth]{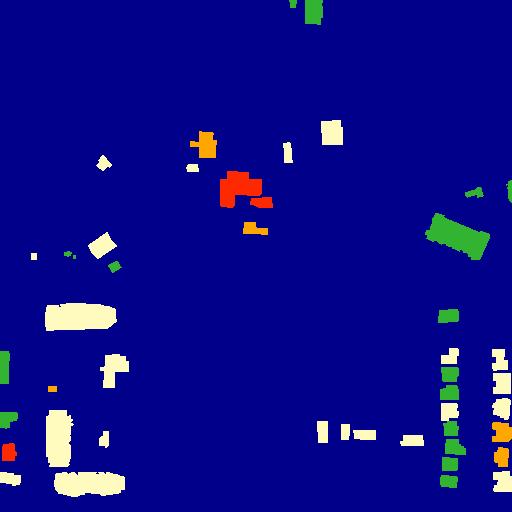}
		\label{fig:1} } \hspace{-2.5mm}  
	\subfloat{
		\includegraphics[width=0.130\linewidth]{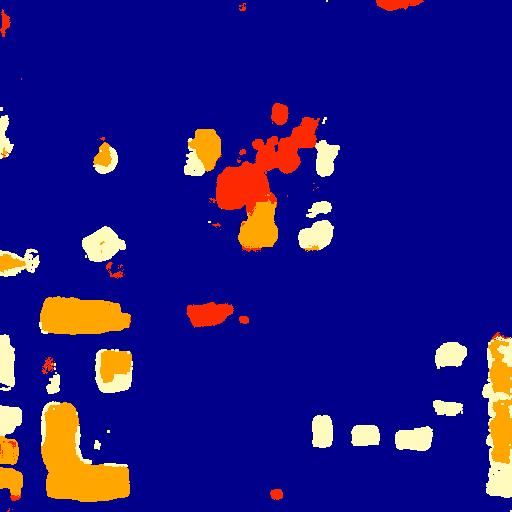}
	     \label{fig:2}}     \hspace{-2.5mm}    
	 \subfloat{
		\includegraphics[width=0.130\linewidth]{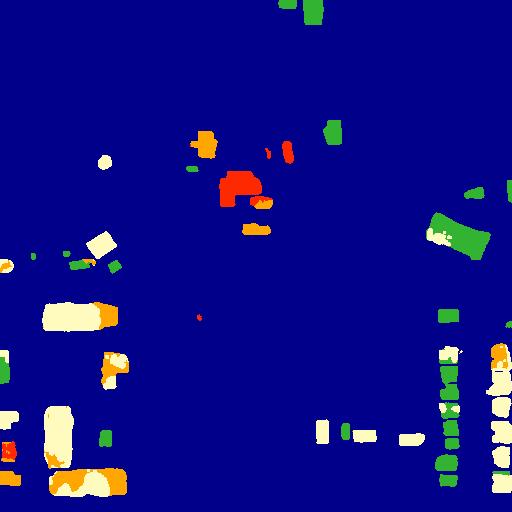}
	     \label{fig:2}}    \hspace{-2.5mm}  
	 \subfloat{
		\includegraphics[width=0.130\linewidth]{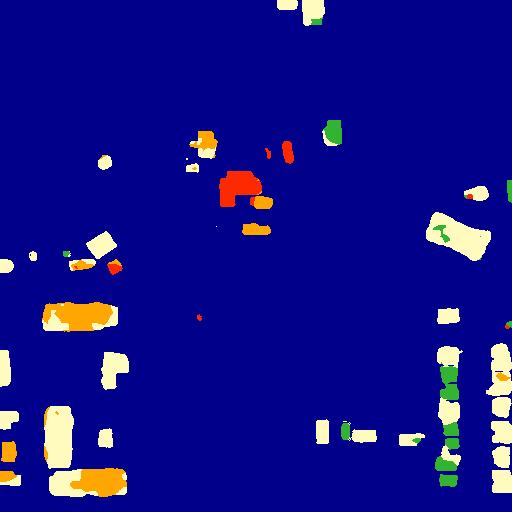}
	     \label{fig:2}}   \hspace{-2.5mm}  
	 \subfloat{
		\includegraphics[width=0.130\linewidth]{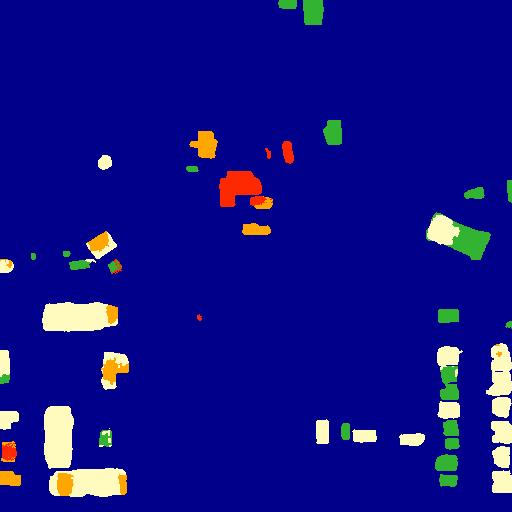}
	     \label{fig:2}}    \\
	 \subfloat{
		\includegraphics[width=0.130\linewidth]{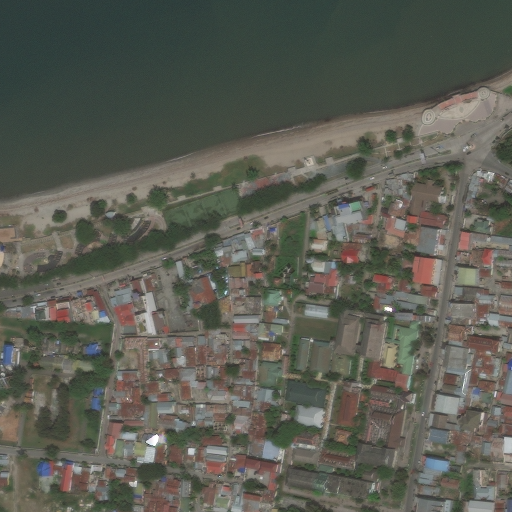}
		\label{fig:1} } \hspace{-2.5mm}  
	\subfloat{
		\includegraphics[width=0.130\linewidth]{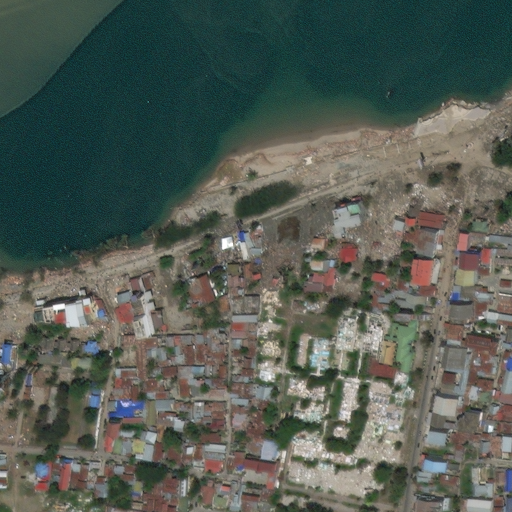}
		\label{fig:2}}   \hspace{-2.5mm}   
	\subfloat{
		\includegraphics[width=0.130\linewidth]{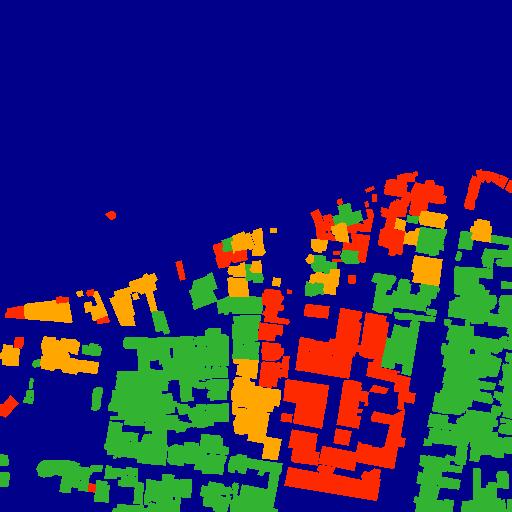}
		\label{fig:1} } \hspace{-2.5mm}  
	\subfloat{
		\includegraphics[width=0.130\linewidth]{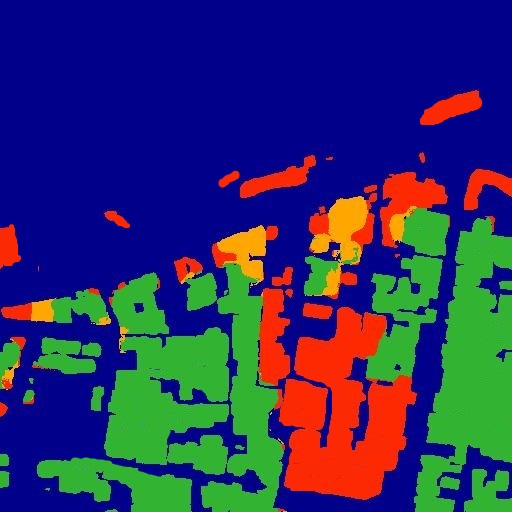}
	     \label{fig:2}}   \hspace{-2.5mm}      
	 \subfloat{
		\includegraphics[width=0.130\linewidth]{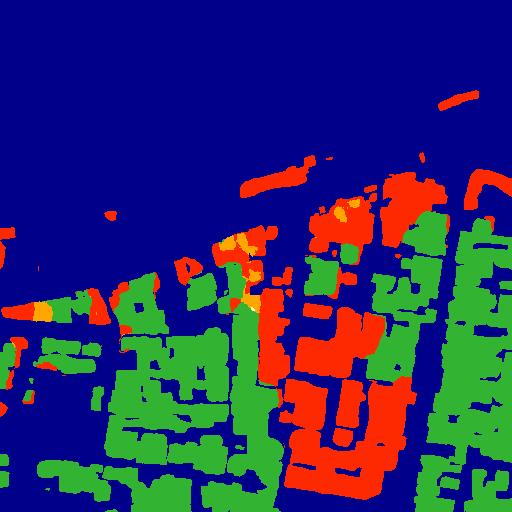}
	     \label{fig:2}}    \hspace{-2.5mm}  
	 \subfloat{
		\includegraphics[width=0.130\linewidth]{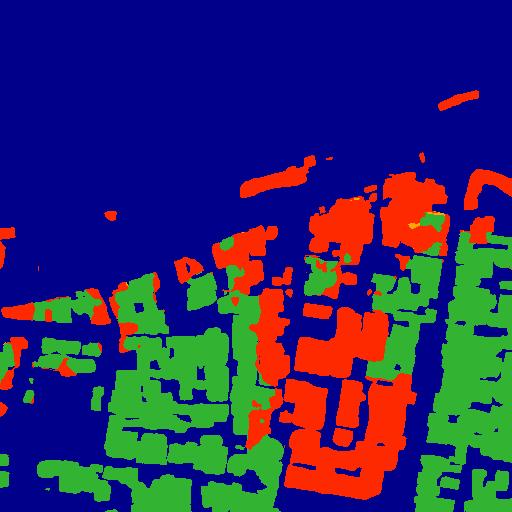}
	     \label{fig:2}}   \hspace{-2.5mm}  
	 \subfloat{
		\includegraphics[width=0.130\linewidth]{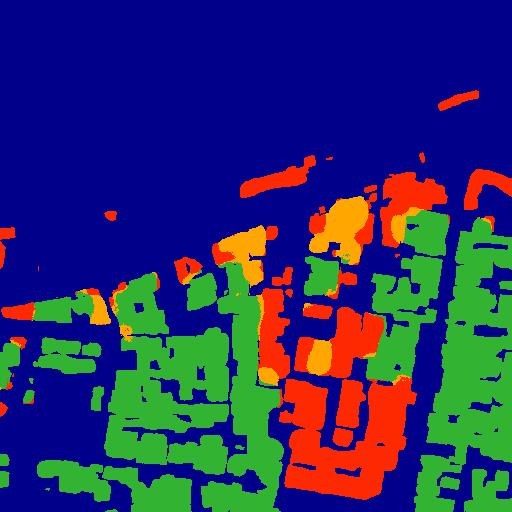}
	     \label{fig:2}}    \\
	 \subfloat{
		\includegraphics[width=0.130\linewidth]{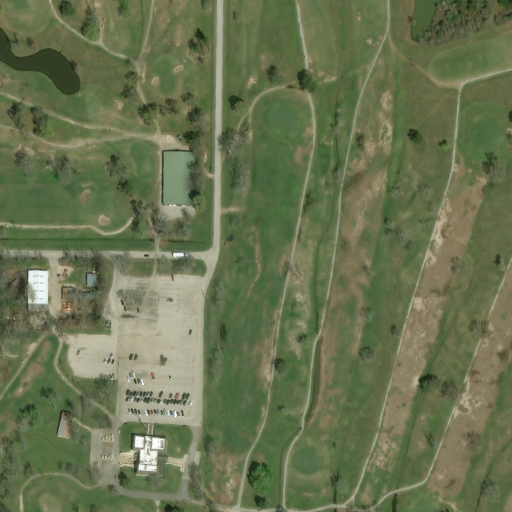}
		\label{fig:1} } \hspace{-2.5mm}  
	\subfloat{
		\includegraphics[width=0.130\linewidth]{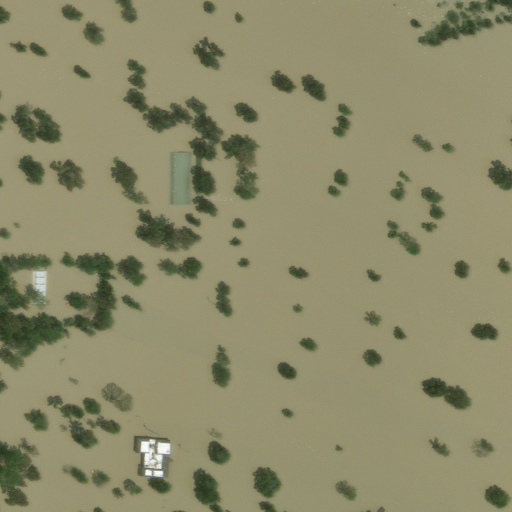}
		\label{fig:2}}  \hspace{-2.5mm}    
	\subfloat{
		\includegraphics[width=0.130\linewidth]{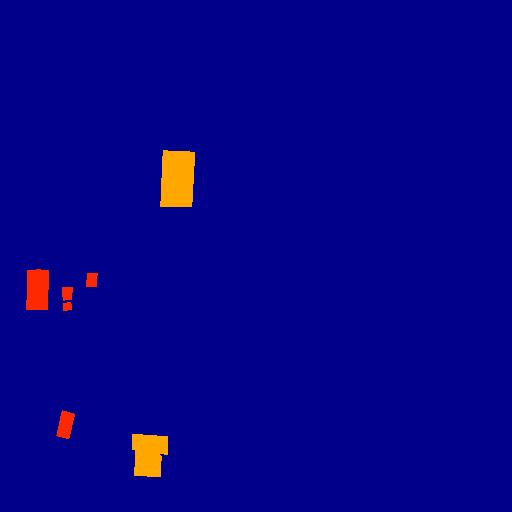}
		\label{fig:1} } \hspace{-2.5mm}  
	\subfloat{
		\includegraphics[width=0.130\linewidth]{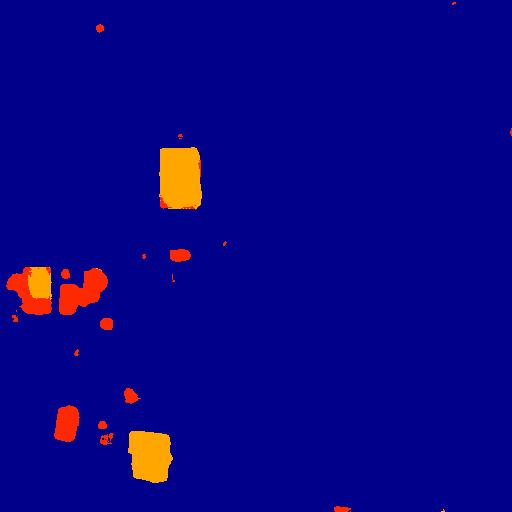}
	     \label{fig:2}}    \hspace{-2.5mm}     
	 \subfloat{
		\includegraphics[width=0.130\linewidth]{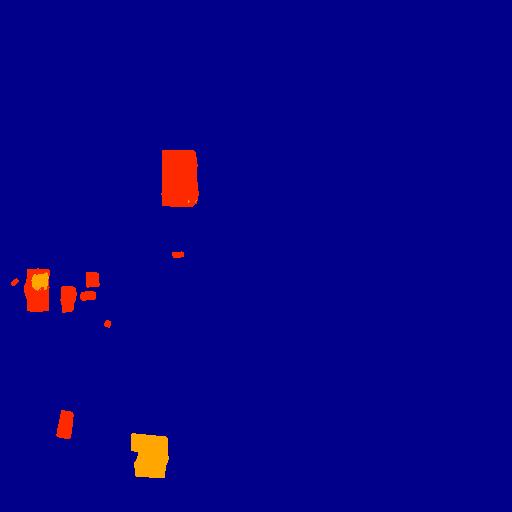}
	     \label{fig:2}}  \hspace{-2.5mm}    
	 \subfloat{
		\includegraphics[width=0.130\linewidth]{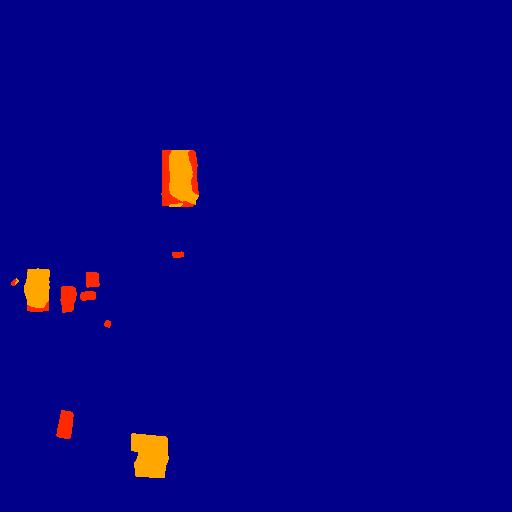}
	     \label{fig:2}}   \hspace{-2.5mm}  
	 \subfloat{
		\includegraphics[width=0.130\linewidth]{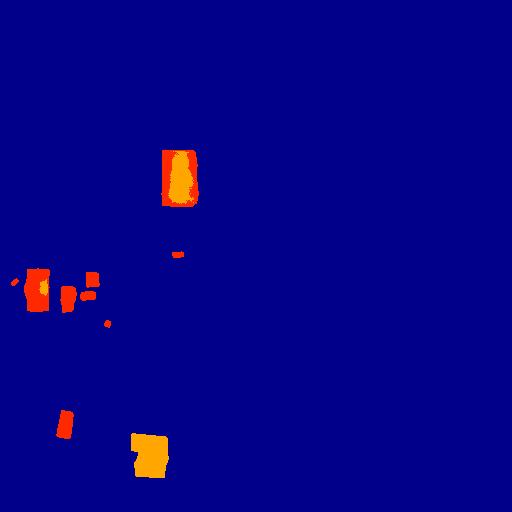}
	     \label{fig:2}}    \\     
	  \subfloat{
		\includegraphics[width=0.130\linewidth]{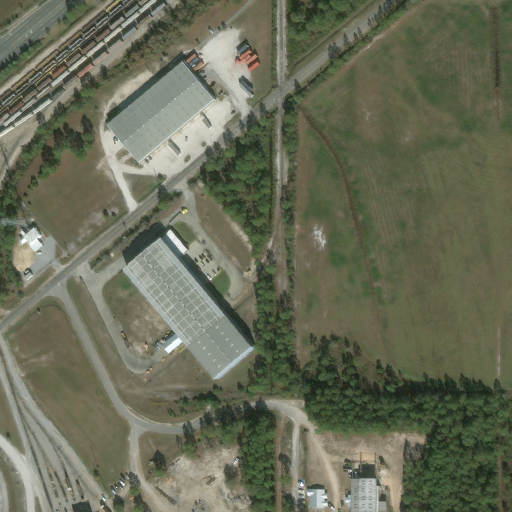}
		\label{fig:1} } \hspace{-2.5mm}  
	\subfloat{
		\includegraphics[width=0.130\linewidth]{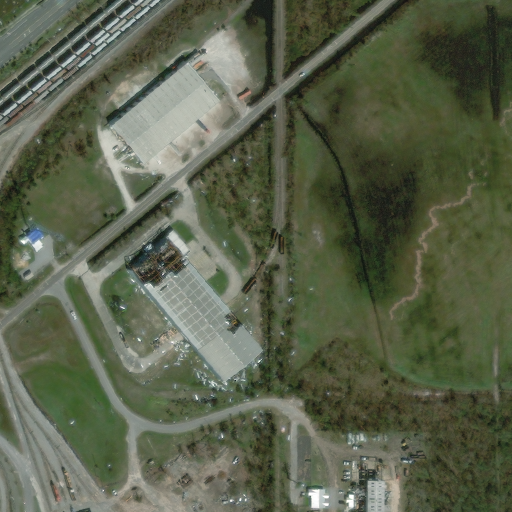}
		\label{fig:2}}    \hspace{-2.5mm}  
	\subfloat{
		\includegraphics[width=0.130\linewidth]{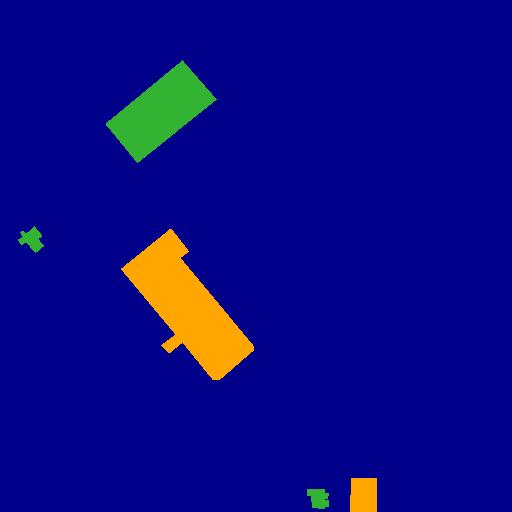}
		\label{fig:1} } \hspace{-2.5mm}  
	\subfloat{
		\includegraphics[width=0.130\linewidth]{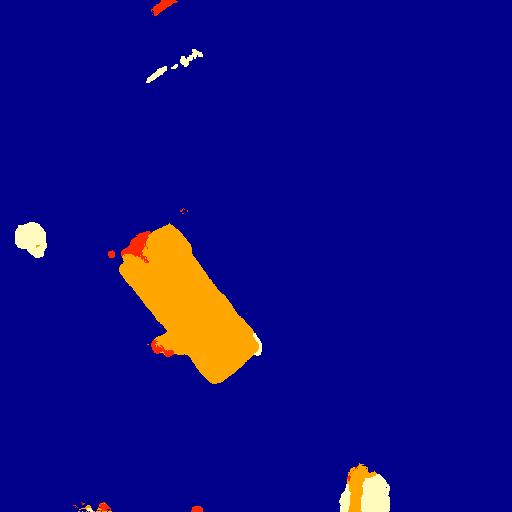}
	     \label{fig:2}}  \hspace{-2.5mm}       
	 \subfloat{
		\includegraphics[width=0.130\linewidth]{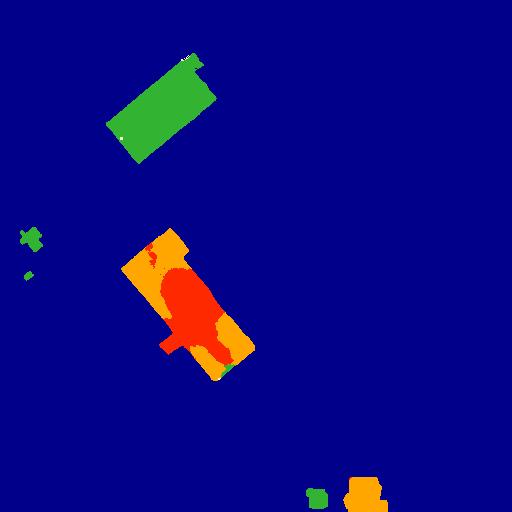}
	     \label{fig:2}}    \hspace{-2.5mm}  
	 \subfloat{
		\includegraphics[width=0.130\linewidth]{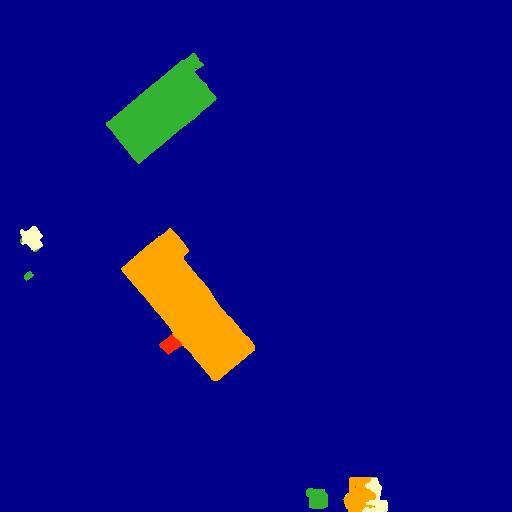}
	     \label{fig:2}}   \hspace{-2.5mm}  
	 \subfloat{
		\includegraphics[width=0.130\linewidth]{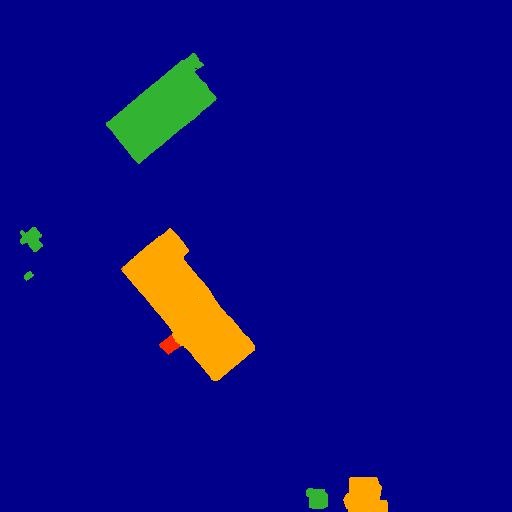}
	     \label{fig:2}}    \\
	  \subfloat{
		\includegraphics[width=0.130\linewidth]{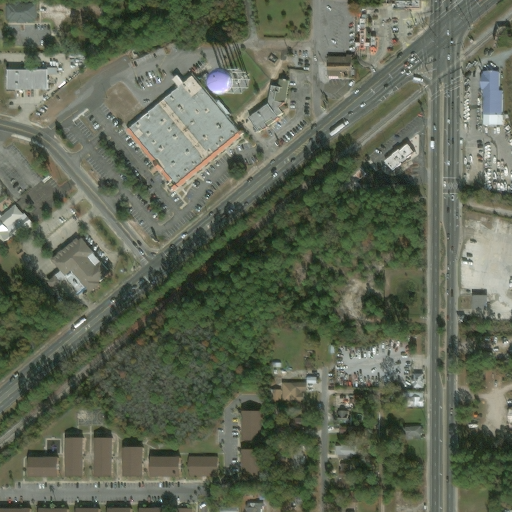}
		\label{fig:1} } \hspace{-2.5mm}  
	\subfloat{
		\includegraphics[width=0.130\linewidth]{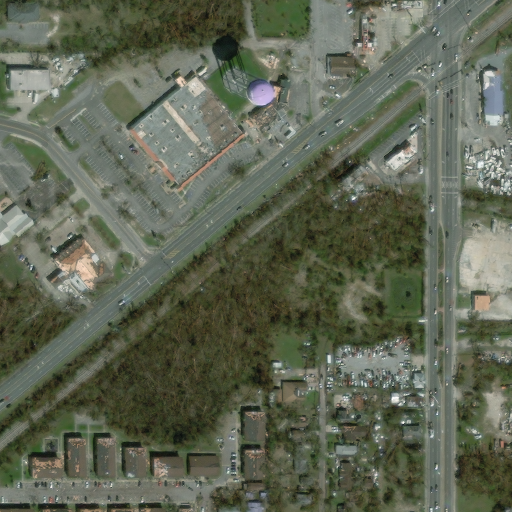}
		\label{fig:2}}  \hspace{-2.5mm}    
	\subfloat{
		\includegraphics[width=0.130\linewidth]{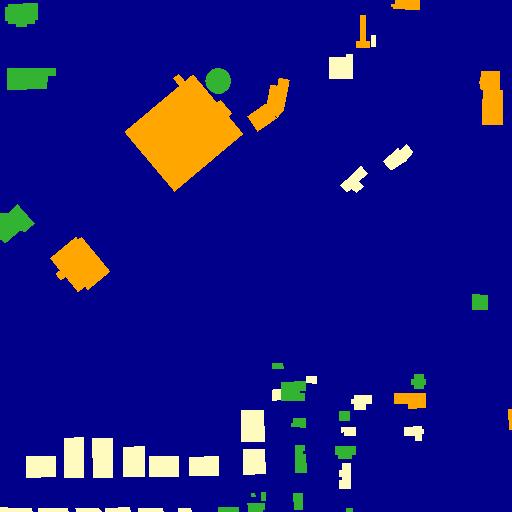}
		\label{fig:1} } \hspace{-2.5mm}  
	\subfloat{
		\includegraphics[width=0.130\linewidth]{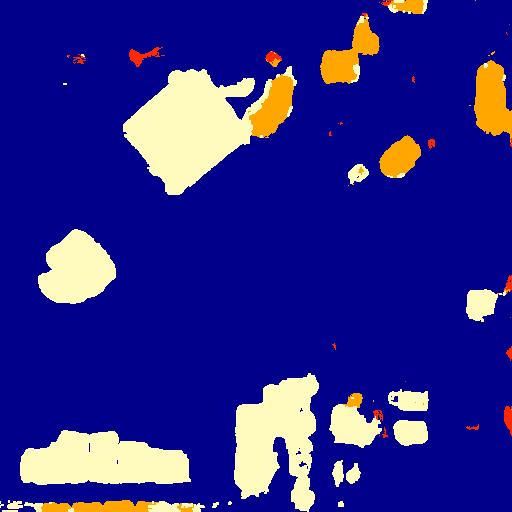}
	     \label{fig:2}}   \hspace{-2.5mm}      
	 \subfloat{
		\includegraphics[width=0.130\linewidth]{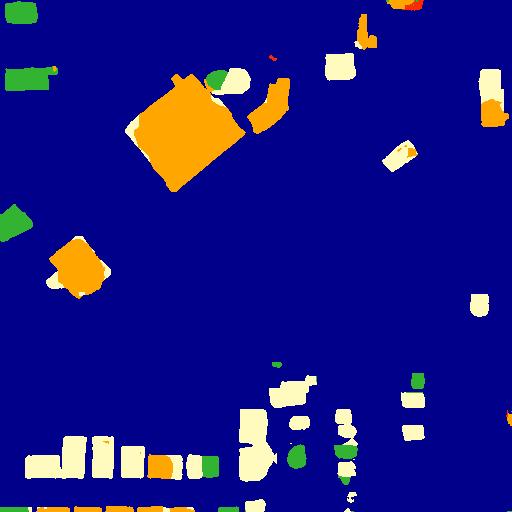}
	     \label{fig:2}}    \hspace{-2.5mm}  
	 \subfloat{
		\includegraphics[width=0.130\linewidth]{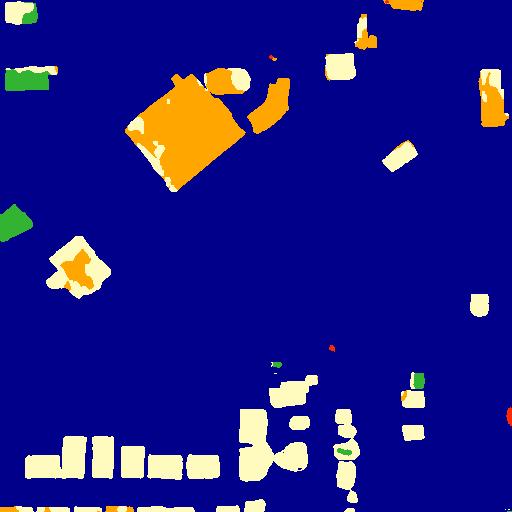}
	     \label{fig:2}}   \hspace{-2.5mm}  
	 \subfloat{
		\includegraphics[width=0.130\linewidth]{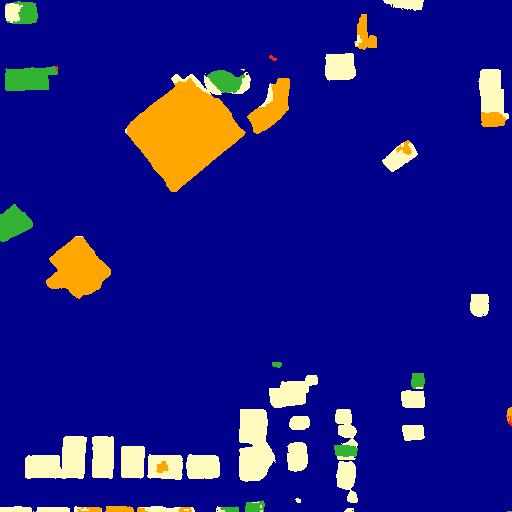}
	     \label{fig:2}}    \\   
	 \subfloat{
		\includegraphics[width=0.130\linewidth]{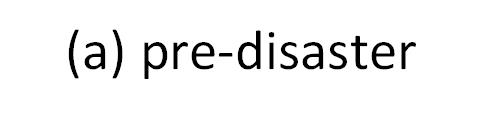}
		\label{fig:1} } \hspace{-2.5mm}  
	\subfloat{
		\includegraphics[width=0.130\linewidth]{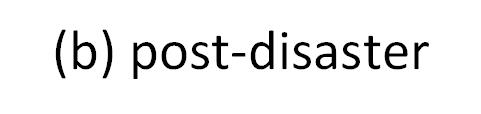}
		\label{fig:2}}    \hspace{-2.5mm}  
	\subfloat{
		\includegraphics[width=0.130\linewidth]{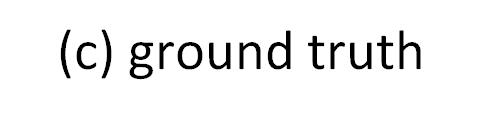}
		\label{fig:1} } \hspace{-2.5mm}  
	\subfloat{
		\includegraphics[width=0.130\linewidth]{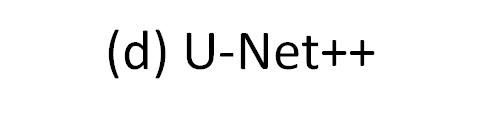}
	     \label{fig:2}}   \hspace{-2.5mm}      
	 \subfloat{
		\includegraphics[width=0.130\linewidth]{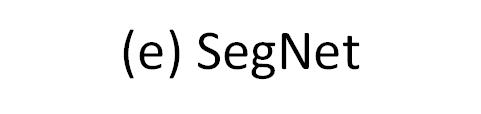}
	     \label{fig:2}}   \hspace{-2.5mm}   
	 \subfloat{
		\includegraphics[width=0.130\linewidth]{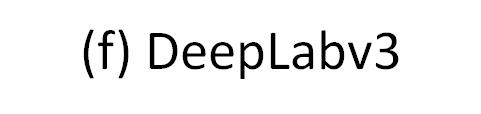}
	     \label{fig:2}}  \hspace{-2.5mm}   
	 \subfloat{
		\includegraphics[width=0.130\linewidth]{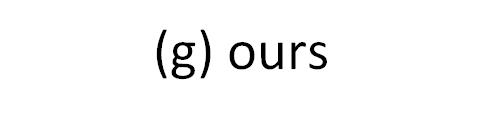}
	     \label{fig:2}}    \\  
	 	 \subfloat{
	\includegraphics[width=0.62\linewidth]{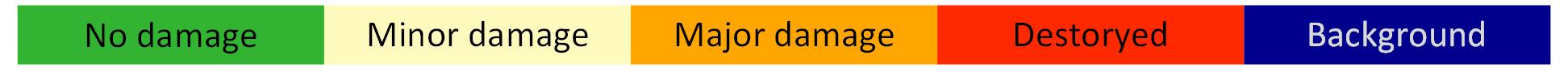}
	     \label{fig:2}} 
	\caption{Damage assessment results. From left to right: (a) pre-disaster image, (b) post-disaster image, (c) the ground-truth, (d) FCN, (e) SegNet, (f) DeepLab and (g) our proposed network.}
	\label{figImgResults}
\end{figure*}

\begin{figure}[t]
	\centering
	\subfloat[1$\times$ stream]{
		\includegraphics[width=0.94\linewidth]{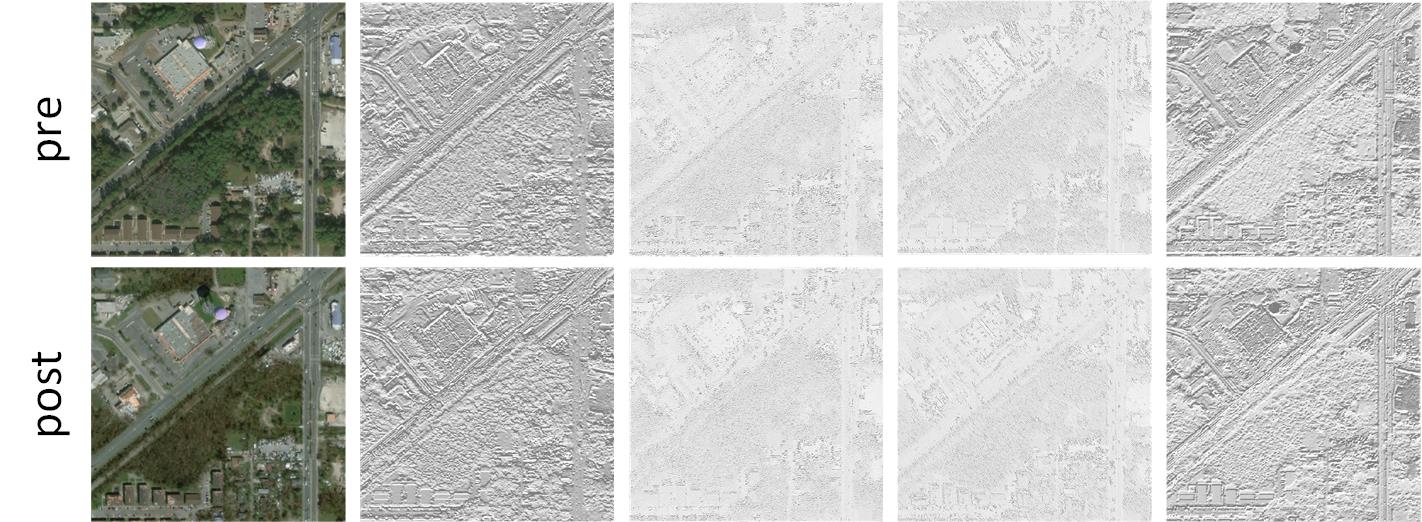}
		\label{fig:1} }  \\
	\subfloat[0.5$\times$ stream]{
		\includegraphics[width=0.94\linewidth]{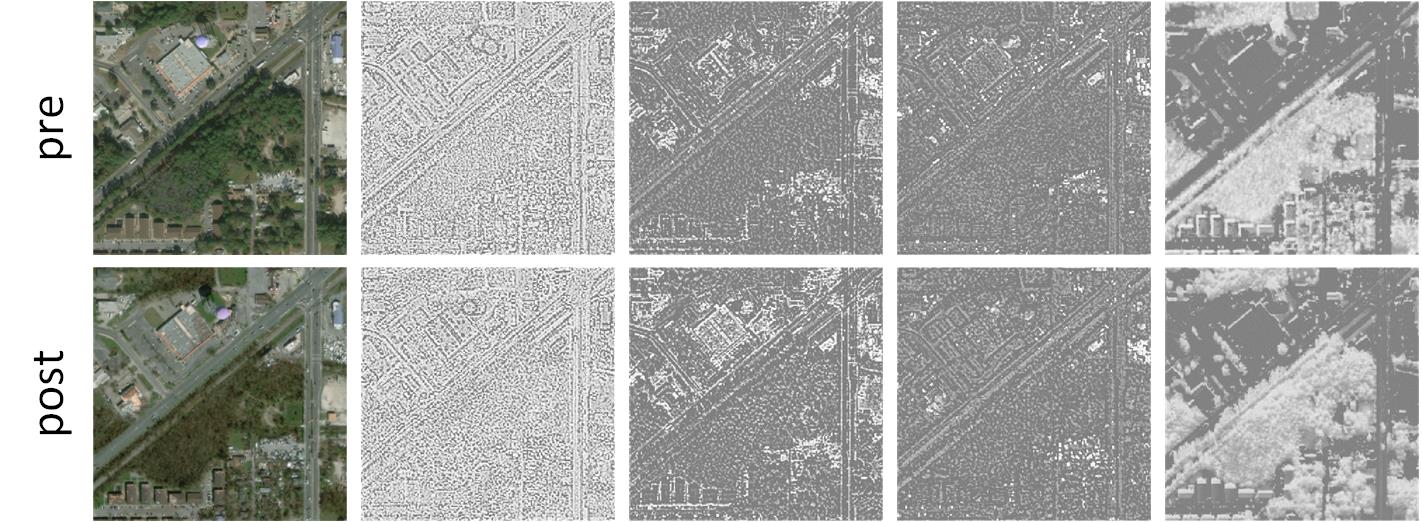}
		\label{fig:2}}   \\
	\subfloat[0.25$\times$ stream]{
		\includegraphics[width=0.94\linewidth]{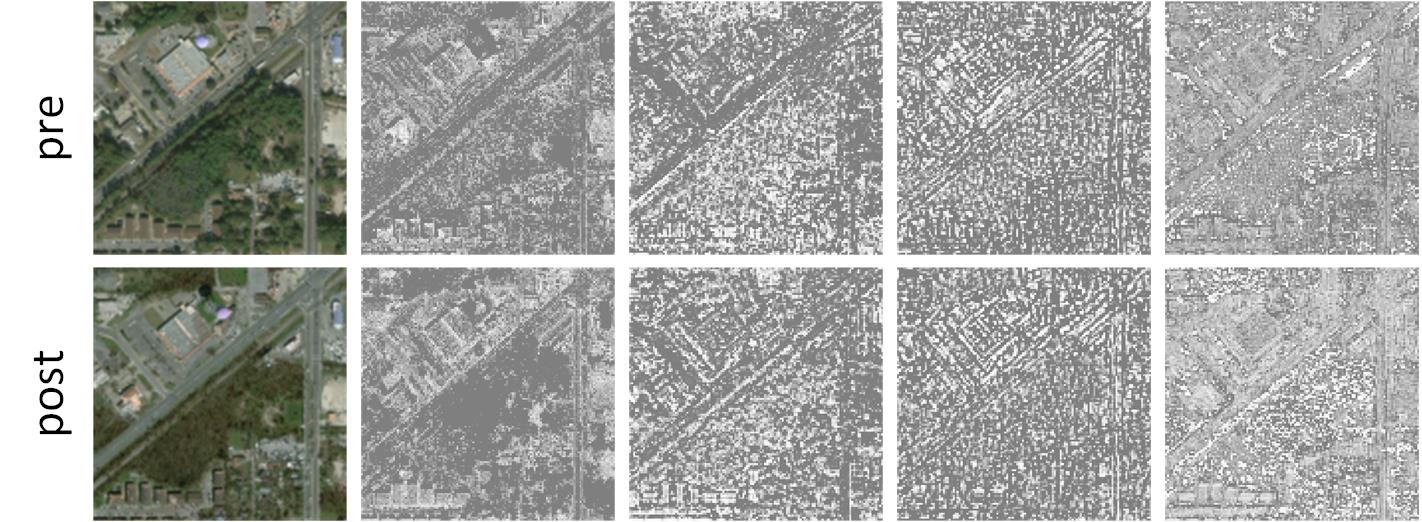}
		\label{fig:1} }   
	\caption{Visualization of feature maps of MFF module. First column: the input images with 1$\times$, 0.5$\times$ and 0.25$\times$ resolution. (a) Feature maps of 1$\times$ stream. (b) Feature maps of 0.5$\times$ stream. (c) Features maps of 0.25$\times$ stream.}
	\label{figMscale}
\end{figure}

\begin{table*}[!t]
	\centering
	\caption{Ablation study of different components (MFF, CDA and CutMix strategy) in the proposed framework. }
	\begin{tabular}{c|ccccccc}
		\hline
		\hline
    \textbf{Method}     & $F1_s$    & $F1_b$ & $F1_d$ & No damage & Minor & Major & Destroyed \\
    \hline
     Ours (vanilla) & 0.789 & 0.864 &	0.757 &	0.923 &	0.578 &	0.760 &	0.869 \\
     vanilla + MFF   & 0.794  &  0.864    & 0.764  & 0.924 & 0.588 & 0.766 &0.870\\
     vanilla + CDA   & 0.795  &  0.864    & 0.765  & 0.923 & 0.592 & 0.769 & 0.871 \\
     vanilla + CutMix  & \textbf{0.802}  &  0.864    & 0.775  & 0.926 & 0.605 & 0.779 & 0.872 \\
		\hline  \hline
	\end{tabular}
\label{tabAblitation}
\end{table*}

\begin{table}[!t]
	\centering
	\caption{Analysis of the MFF module. The combinations of feature streams with different scales (1$\times$, 0.5$\times$ and 0.25$\times$) are used for comparison. }
	\setlength{\tabcolsep}{1.15mm} {
	\begin{tabular}{c|cccc|c}
	\hline \hline
    Method    & No damage & Minor & Major & Destroyed  & $F1_s$ \\
    \hline
    Ours $1\times$ &0.923 &0.578 &0.760 &0.869 & 0.789  \\
     $1\times$ $+$ $0.5\times$ &  0.924 & 0.586 &  0.763  & 0.870   & 0.793\\
     $1\times$ $+$ $0.25\times$ & 0.924 &  0.580  & 0.762  & 0.870 &0.791\\
   $1\times$ $+$ $0.5\times$ $+$ $0.5\times$  &   0.924 & 0.588 & 0.766 &0.870    & \textbf{0.794}\\
		\hline  \hline
	\end{tabular}}
\label{tabAbNMFF}
\end{table}

\begin{table}[!t]
	\centering
	\caption{Influence of the positions (dconv1, dconv2 and dconv3) of CDA modules in the network. }
	\setlength{\tabcolsep}{1.2mm} {
	\begin{tabular}{c|ccc|c}
		\hline \hline
		 Method & dconv1 & dconv2 & dconv3  & $F1_s$ \\
		 \hline
    Ours (vanilla) & & & &  0.789  \\
    vanilla + CDA & \checkmark  &  &        & 0.790\\
    vanilla + CDA &  & \checkmark &      & 0.791\\
    vanilla + CDA &  &    &  \checkmark       & 0.793\\
    vanilla + CDA & \checkmark  & \checkmark &    &      0.794\\
    vanilla + CDA & \checkmark  & \checkmark & \checkmark   &   \textbf{0.795}\\
		\hline  \hline
	\end{tabular}}
\label{tabAbCDA}
\end{table}

\begin{table}[!t]
	\centering
	\caption{The damage assessment results of using CDA or SE modules.}
	\setlength{\tabcolsep}{1.15mm} {
	\begin{tabular}{c|cccc|c}
		\hline \hline
		 module & No damage & Minor & Major & Destroyed &  $F1_s$  \\
		 \hline
		 SE cha \cite{royRecalibratingFullyConvolutional2019} & 0.923 & 0.579 & 0.761 & 0.871  & 0.790  \\
		 SE spa  \cite{royRecalibratingFullyConvolutional2019} &  0.923 & 0.580 & 0.762 & 0.871 & 0.792\\
		 SE cha + spa  \cite{royRecalibratingFullyConvolutional2019} & 0.923 & 0.586 & 0.763 & 0.871  &0.793     \\
		 CDA   &   0.923 & 0.592 & 0.769 & 0.871   & \textbf{0.795}          \\
	\hline
	\end{tabular}}
\label{tabAbSE}
\end{table}

\begin{table}[!t]
	\centering
	\caption{Analysis of CutMix applied to different damage levels.}
	\setlength{\tabcolsep}{1.5mm} {
	\begin{tabular}{c|cccc|c}
	\hline \hline
    Method & No damage & Minor & Major & Destroyed  & $F1_s$ \\
        \hline
    Ours (vanilla) & & & & & 0.789  \\
    vanilla + CutMix & \checkmark  & \checkmark &  \checkmark  &   \checkmark  & 0.790\\
    vanilla + CutMix  &  & \checkmark &  \checkmark  &   \checkmark  & 0.797\\
    vanilla + CutMix &   & \checkmark &  \checkmark  &     & \textbf{0.802}\\
    vanilla + CutMix  &   & \checkmark &    &     & 0.795\\
	\hline  \hline
	\end{tabular}}
\label{tabAbCutmix}
\end{table}

\begin{table}[!t]
	\centering
	\caption{Computational cost and inference speed analysis of the network in Stage 2.}
	\setlength{\tabcolsep}{1.15mm} {
	\begin{tabular}{c|ccc}
		\hline \hline
    \textbf{Method}     & Params (M)   & FLOPs (G)  & Inference time (s/img) \\
    \hline
     Ours (vanilla)  & 32.5  & 92.9  &0.174\\
     vanilla + MFF & 33.8 & 141.7  & 0.186\\
     vanilla + CDA    & 33.1  &  107.6  &0.181  \\
     Ours (BDANet) & 34.4 & 155.4 & 0.190\\
	\hline
	\end{tabular}}
\label{tabParam}
\end{table}

Some visual results of building segmentation are presented in Fig. \ref{figBuildingloc}. By decomposing the framework into two stages, the network can focus on building segmentation in Stage 1, achieving good segmentation results as compared with the ground truth segmentation maps. 
Therefore, the results can be used to effectively guide the building locations for damage assessment in Stage 2.

Fig. \ref{figImgResults} shows a visual comparison of damage assessment results of different networks. For U-Net++, it is obvious that this method can not provide precise outlines of building objects and thus many segmentation mistakes can be found. For FCN, more mis-classification can be found from its classification map compared with other methods. We can also observe that much mis-classification come from the levels of minor  and major damage because both two levels are difficult classes. Compared with FCN and SegNet, DeepLabv3 provides a better visual result. Overall, our proposed BDANet achieves the best performance with less damage assessment errors.

\subsection{Ablation Study}
To investigate the contribution of MFF module, CDA module and CutMix data augmentation, we carry out ablation studies on these components. Table \ref{tabAblitation} reports the results of vanilla network with these individual components. It can be seen that CDA and MFF bring about 0.6\% and 0.5\% improvement in $F1_s$, respectively.  Moreover, the CutMix strategy on difficult classes boosts the overall performance ($F1_s$) by 1.3\%, which is a considerable gain. In particular, the accuracy of \textit{minor damage} class is improved by 2.7\% (0.578 vs 0.605) with CutMix. 

1) \textit{Analysis of MFF.} As illustrated in Fig. \ref{figFramework}, three feature streams are used to extract the features from different image resolutions. Therefore, this ablation study investigates the effectiveness of the three streams. Note that when only the stream with the original image resolution (1$\times$) is used, the framework is consistent with the vanilla network. Table \ref{tabAbNMFF} reports the results of different streams applied in the network, where the $0.5\times$ and $0.25\times$ scales yield about 0.4\% and 0.2\% performance gain in $F1_s$,
respectively. When three streams are used together in the network, an improvement of 0.5\% is obtained, indicating that learning features from different image scales is useful for improving the overall performance of damage assessment.

\begin{figure}[t]
	\centering
	\subfloat[]{
		\includegraphics[width=0.82\linewidth]{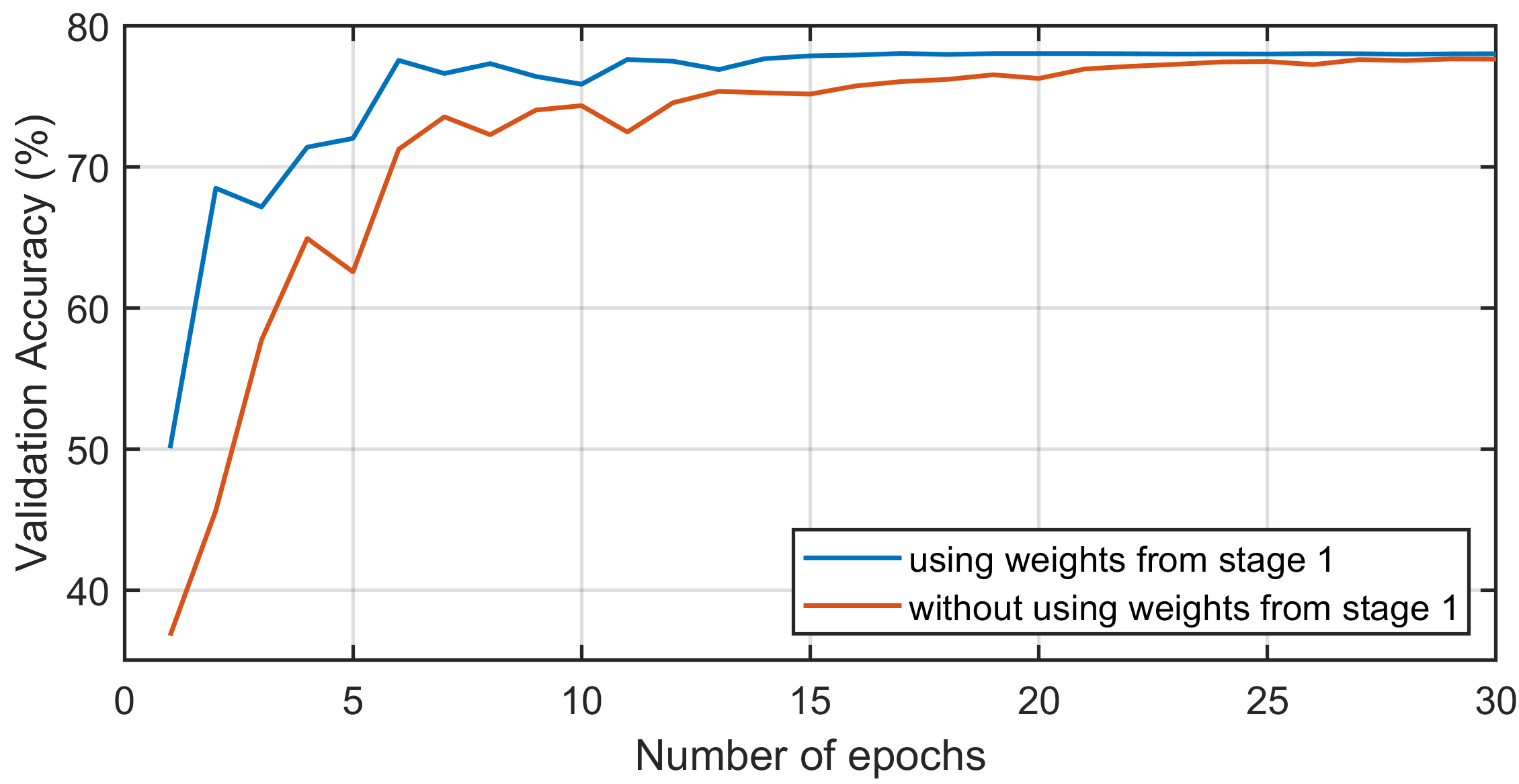}
		\label{fig:1} }           \\
	\subfloat[]{
		\includegraphics[width=0.82\linewidth]{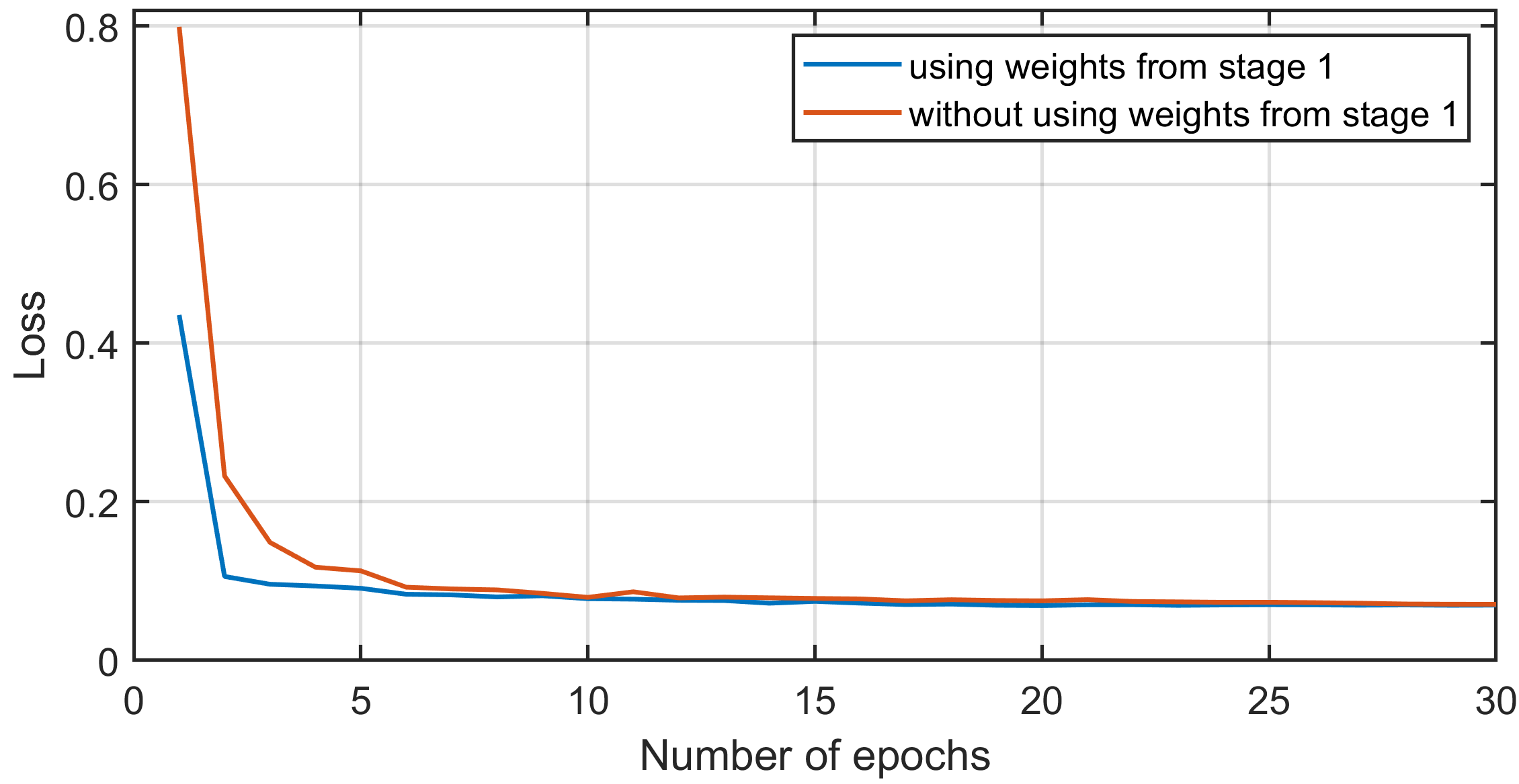}
		\label{fig:2}}    
	\caption{Evaluation of the training efficiency in Stage 2 by applying network weights from Stage 1 as initialization. (a) Validation accuracy (\%). (b) Cross-entropy loss. }
	\label{figCurve}
\end{figure}

The feature maps from different streams are also depicted in Fig. \ref{figMscale}. Specifically, we visualize several feature maps extracted from the first convolutional layer of the three streams.  We can observe that local and detailed information can be extracted from the 1$\times$ stream, while more global information is captured from streams with lower resolution (0.5$\times$ and 0.25$\times$).

\begin{table}[t]
	\centering
	\caption{Comparison of different methods on LEVIR-CD dataset.}
	\setlength{\tabcolsep}{2.6mm} {
		\begin{tabular}{r|ccc}
			\hline \hline
		Method & \textit{precision} & \textit{recall} & \textit{F1} \\
		\hline
	     WNet \cite{houWNetCDGANBitemporal2020} &0.879 &	 0.995 &	0.934  \\
        U-Net++ \cite{pengEndtoEndChangeDetection2019} & 0.890 &	0.996	&0.940  \\
	BDANet (Stage 2) & 0.895 	& 0.995	& \textbf{0.942}  \\
\hline	\hline	
	\end{tabular}}
	\label{tabCD}
\end{table}

\begin{figure*}[t]
	\centering
	\subfloat{
		\includegraphics[width=0.130\linewidth]{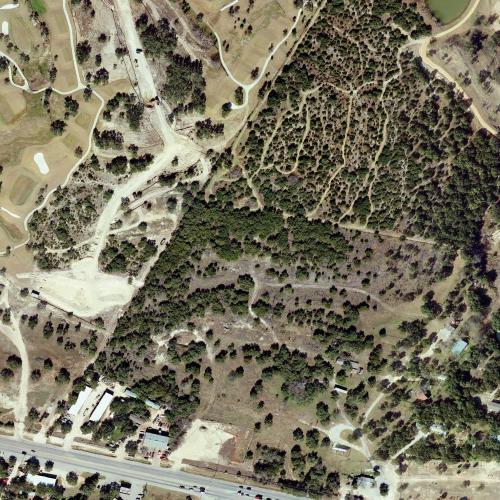}
		\label{fig:1} } \hspace{-2.5mm}  
	\subfloat{
		\includegraphics[width=0.130\linewidth]{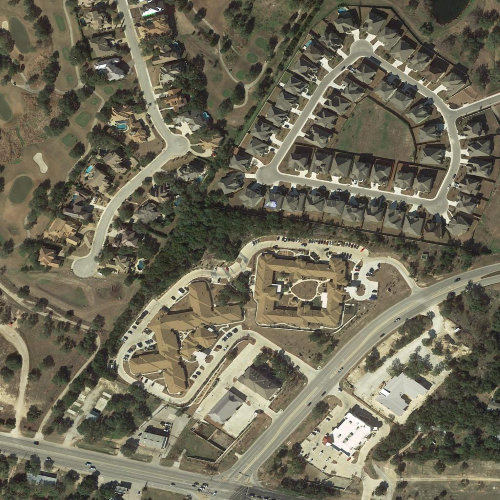}
		\label{fig:2}}    \hspace{-2.5mm}  
	\subfloat{
		\includegraphics[width=0.130\linewidth]{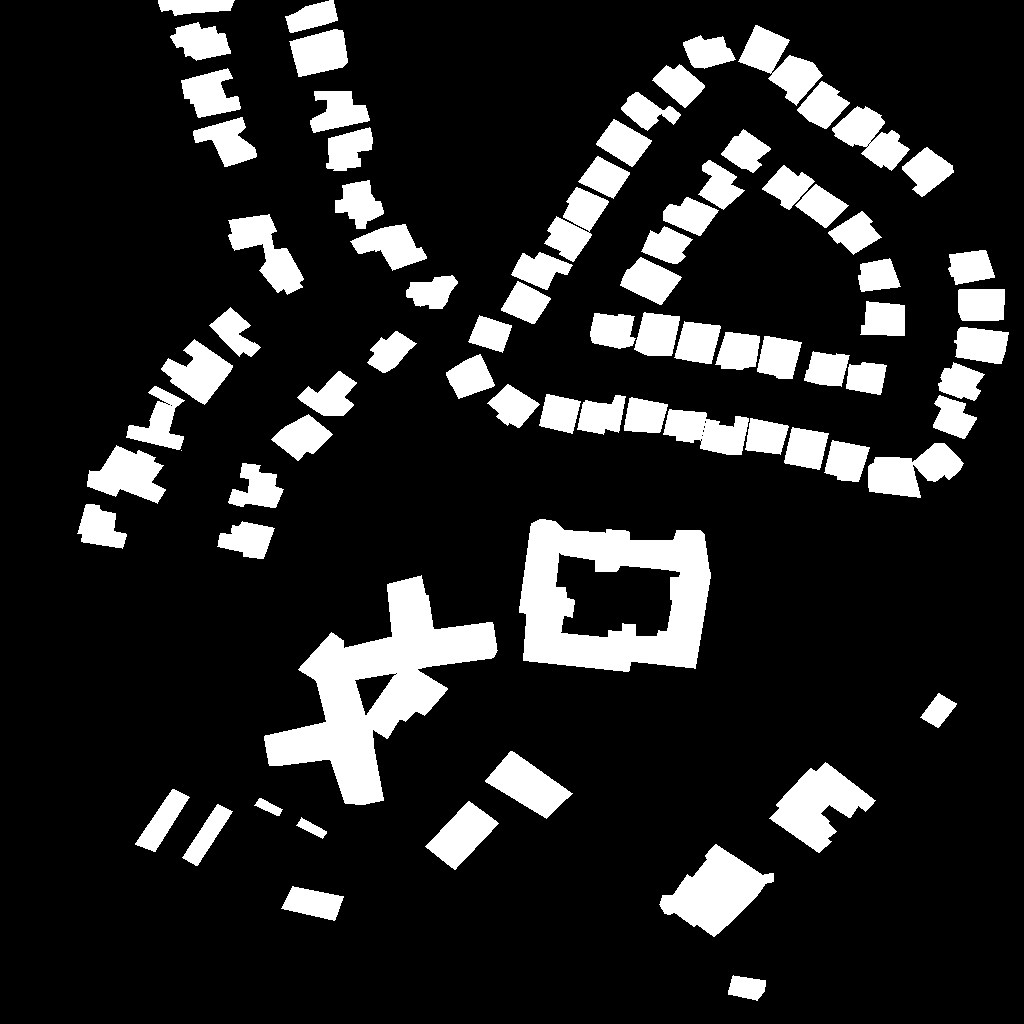}
		\label{fig:1} } \hspace{-2.5mm}  
	\subfloat{
		\includegraphics[width=0.130\linewidth]{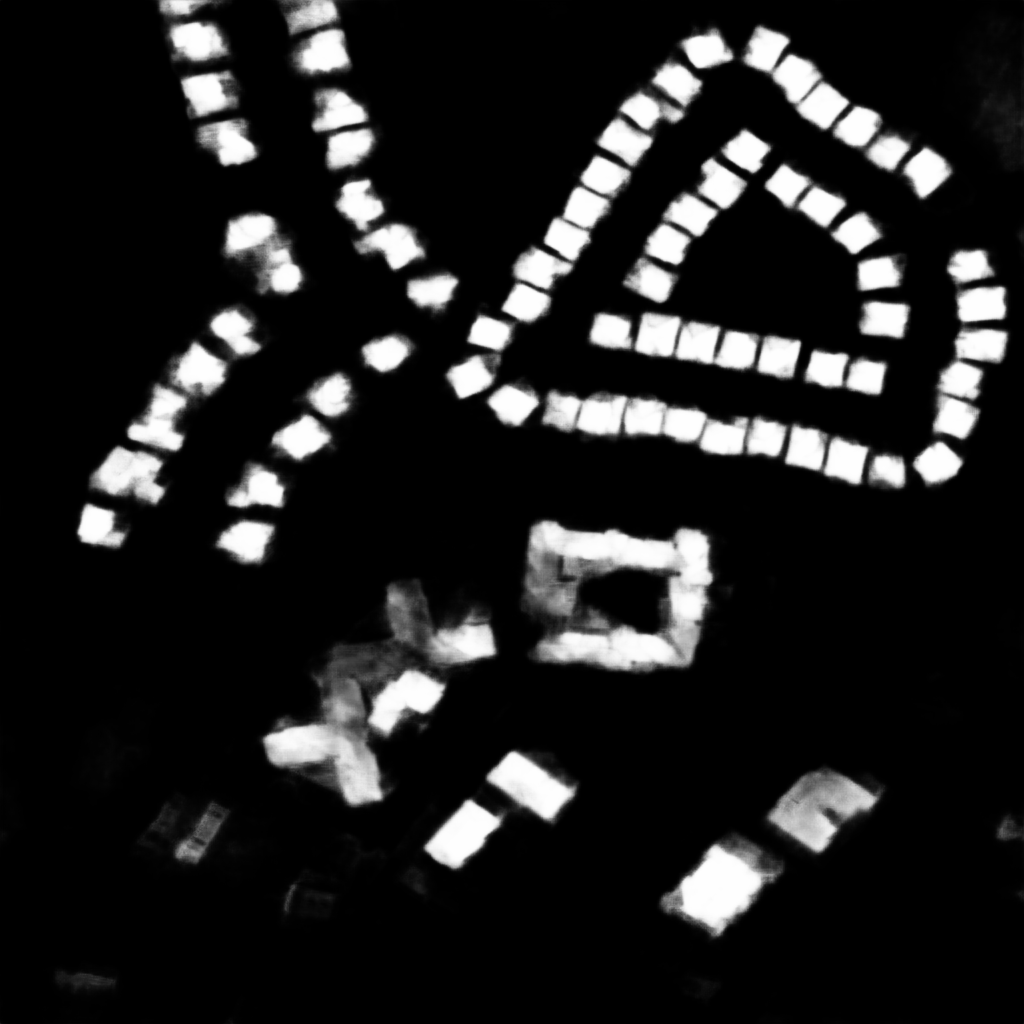}
	     \label{fig:2}}     \hspace{-2.5mm}    
	 \subfloat{
		\includegraphics[width=0.130\linewidth]{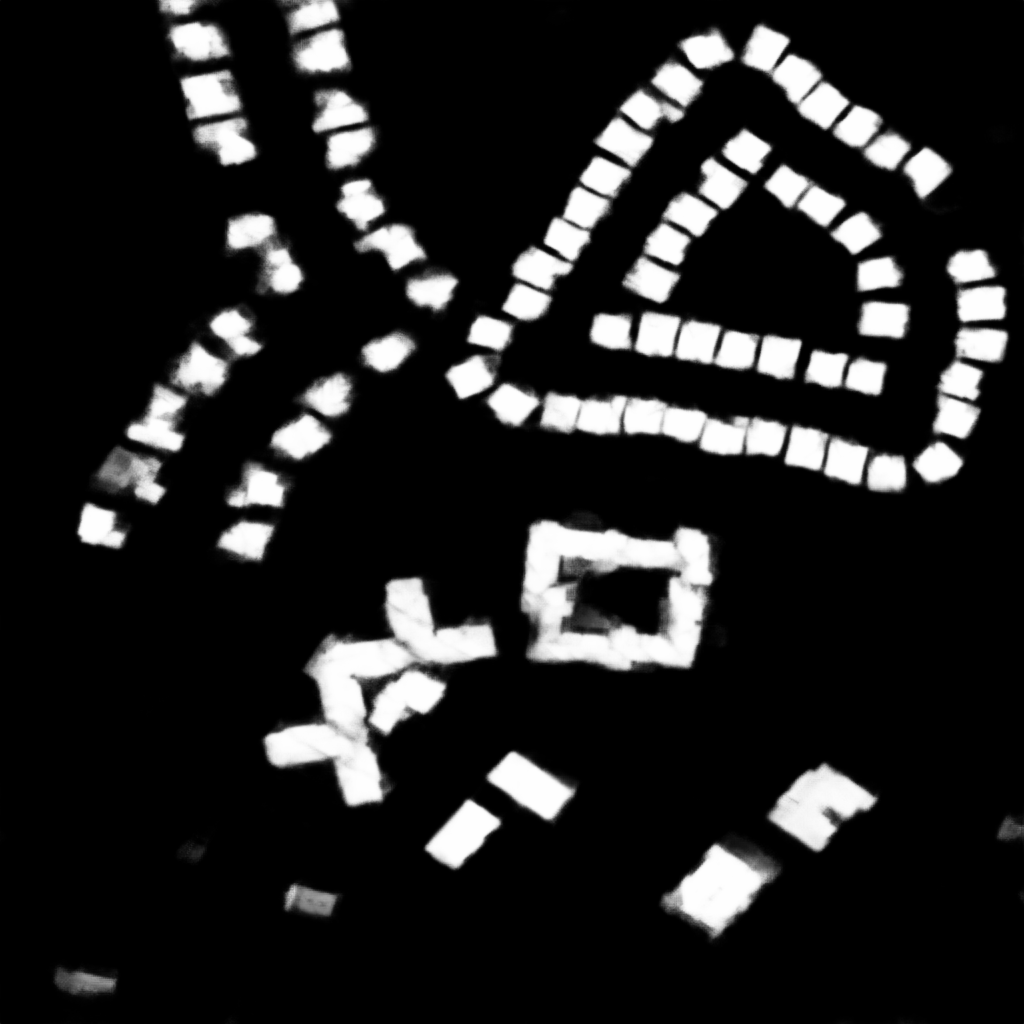}
	     \label{fig:2}}    \hspace{-2.5mm}  
	 \subfloat{
		\includegraphics[width=0.130\linewidth]{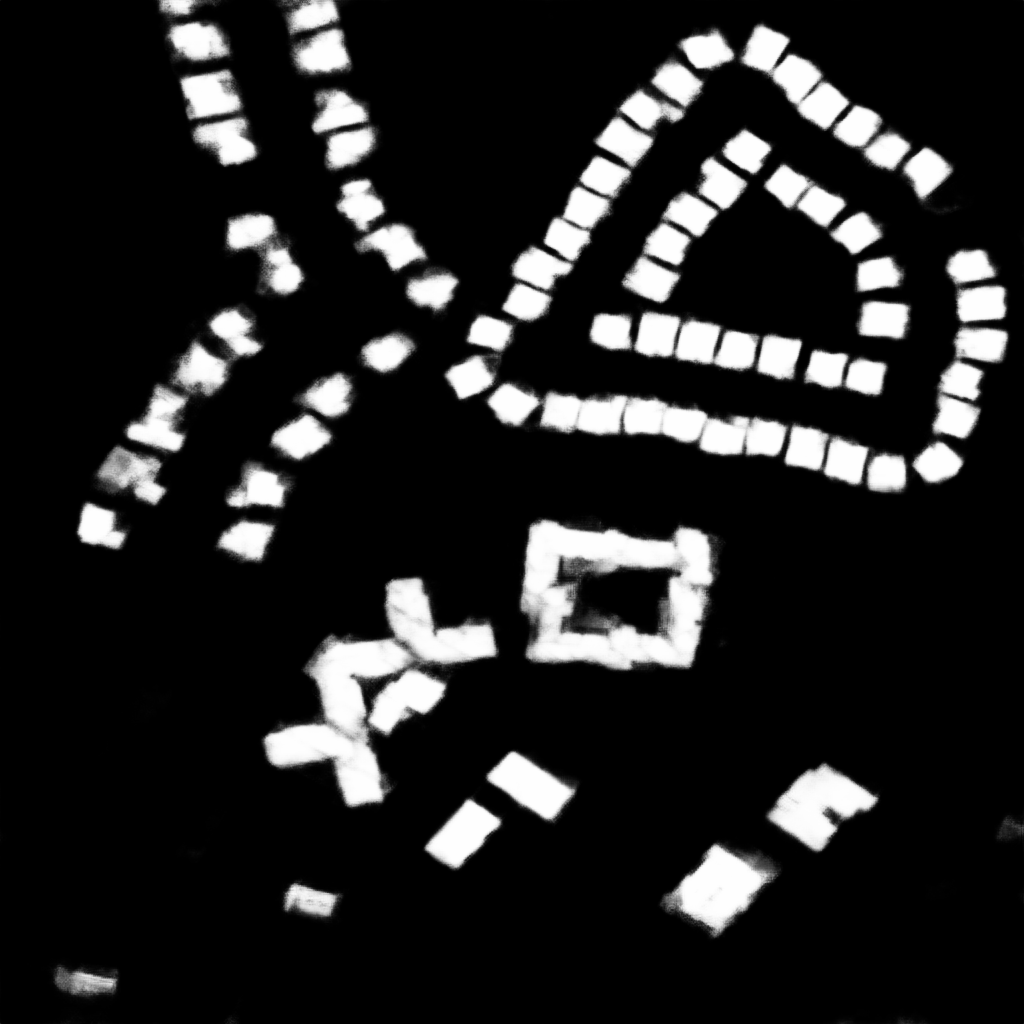}
	     \label{fig:2}}   \hspace{-2.5mm}    \\
	 \subfloat{
		\includegraphics[width=0.130\linewidth]{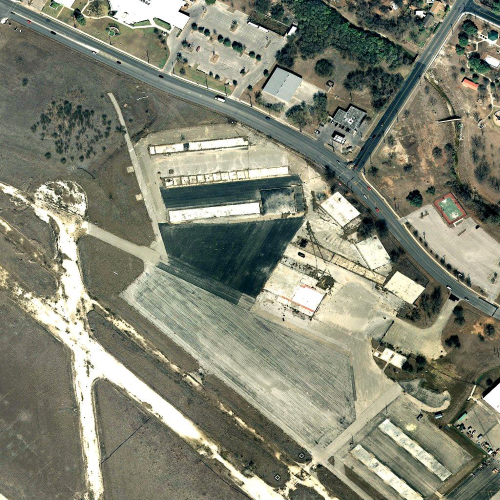}
		\label{fig:1} } \hspace{-2.5mm}  
	\subfloat{
		\includegraphics[width=0.130\linewidth]{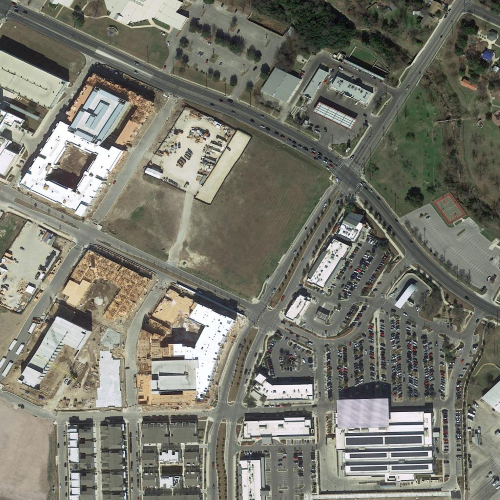}
		\label{fig:2}}  \hspace{-2.5mm}    
	\subfloat{
		\includegraphics[width=0.130\linewidth]{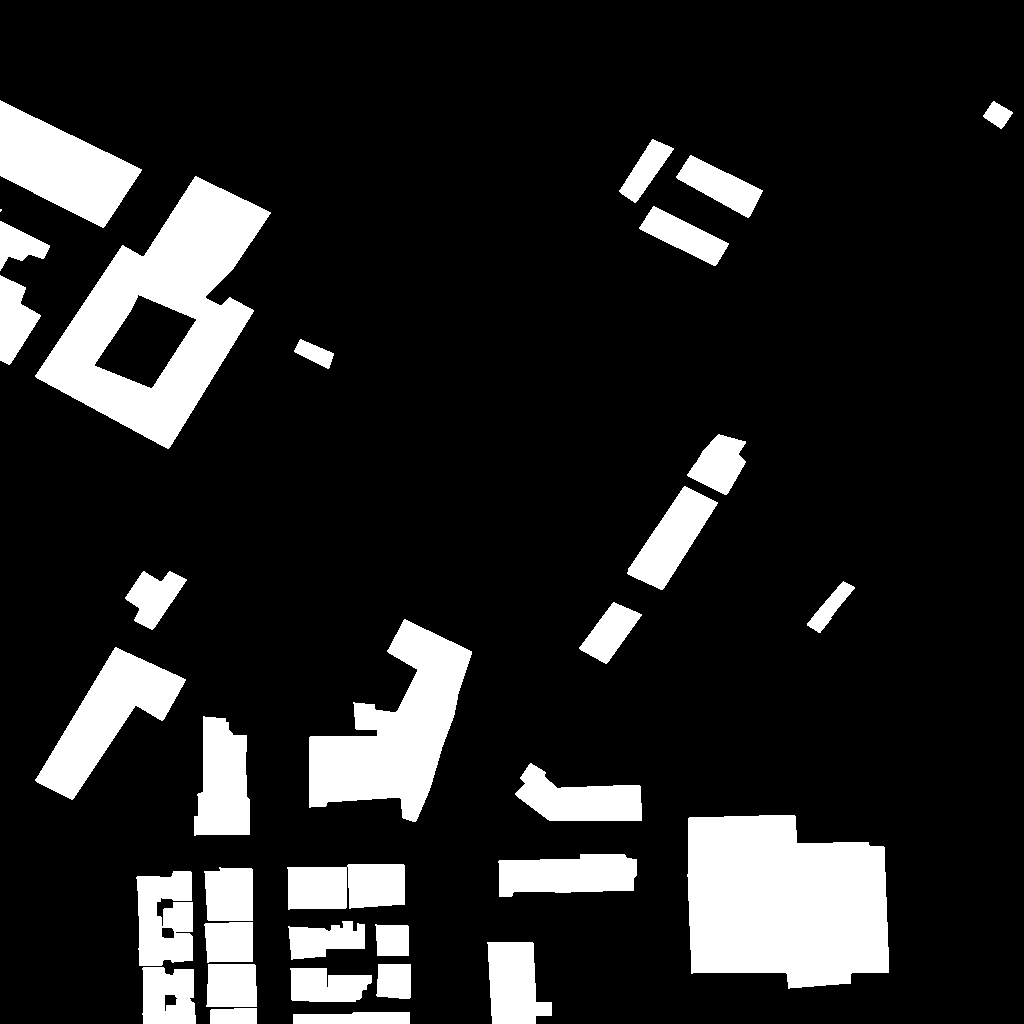}
		\label{fig:1} } \hspace{-2.5mm}  
	\subfloat{
		\includegraphics[width=0.130\linewidth]{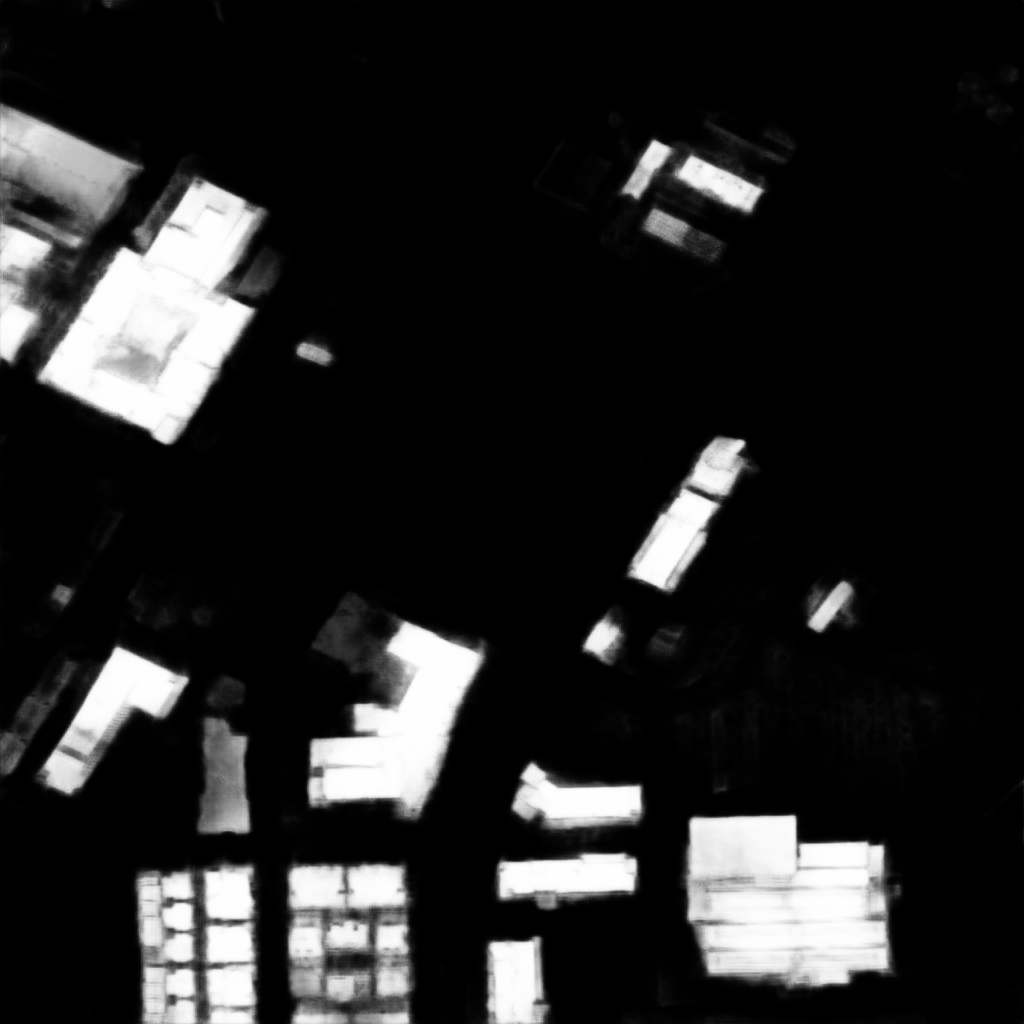}
	     \label{fig:2}}    \hspace{-2.5mm}     
	 \subfloat{
		\includegraphics[width=0.130\linewidth]{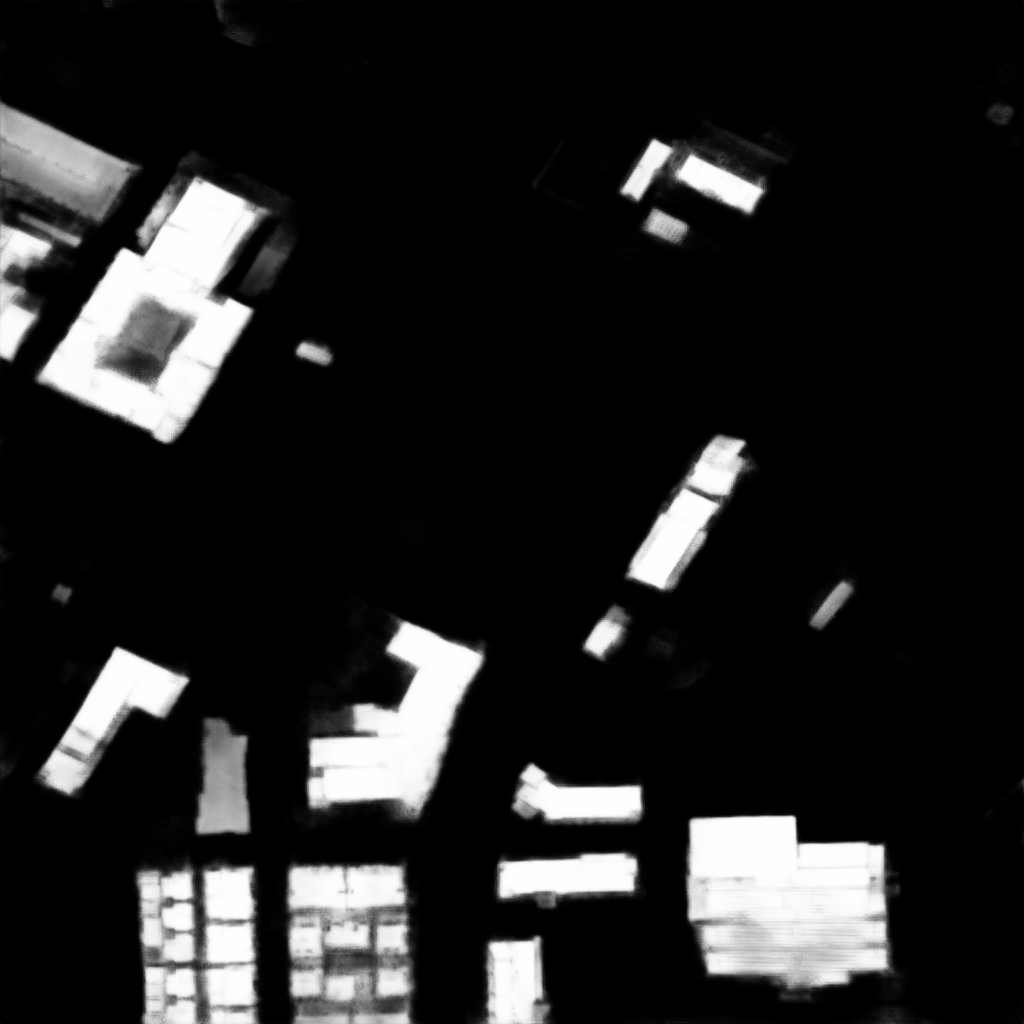}
	     \label{fig:2}}  \hspace{-2.5mm}    
	 \subfloat{
		\includegraphics[width=0.130\linewidth]{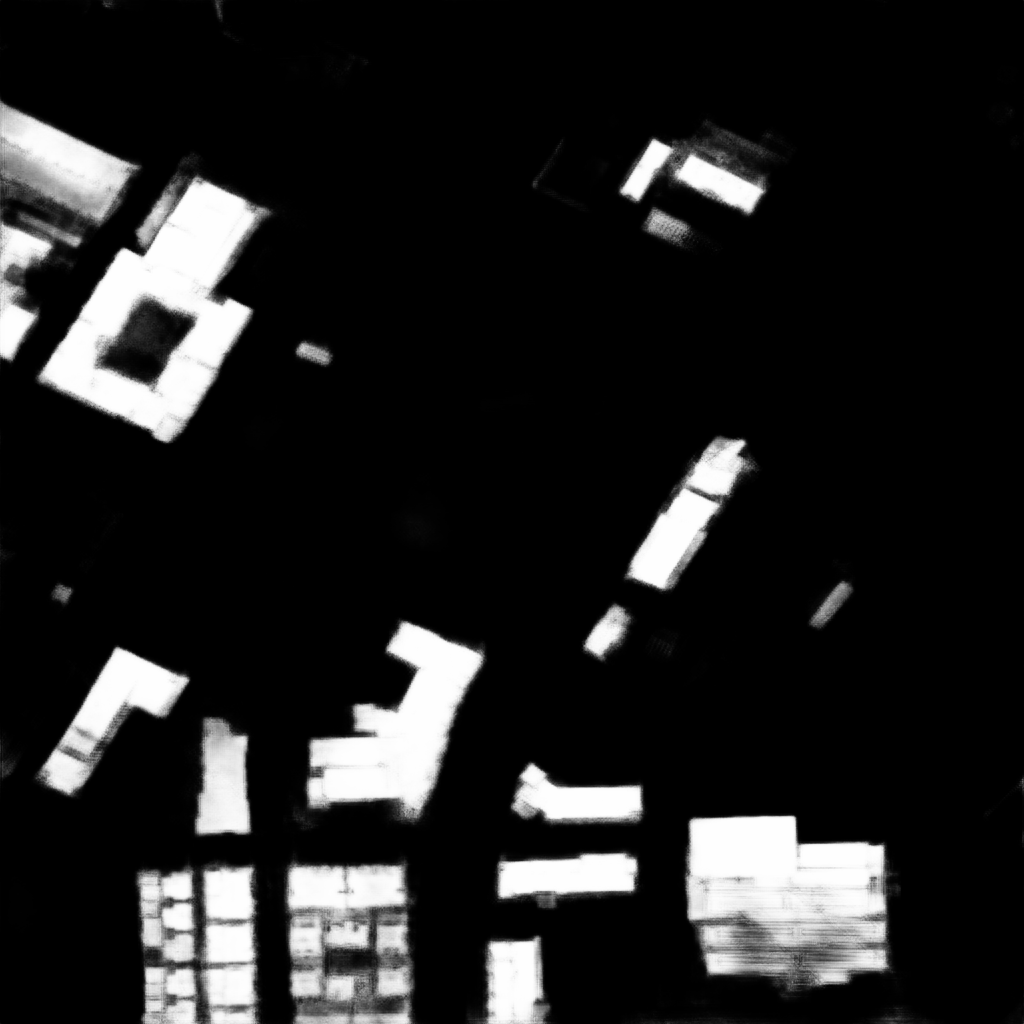}
	     \label{fig:2}}   \hspace{-2.5mm}     \\     
	  \subfloat{
		\includegraphics[width=0.130\linewidth]{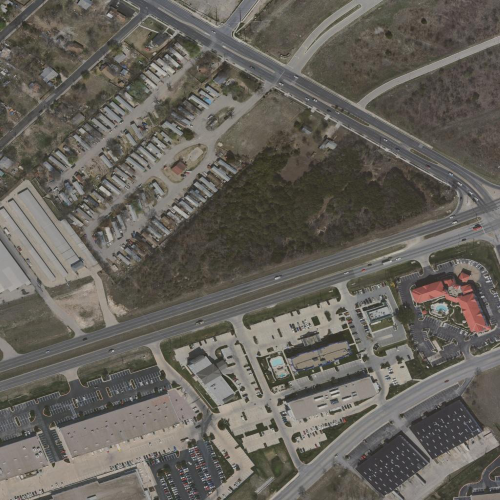}
		\label{fig:1} } \hspace{-2.5mm}  
	\subfloat{
		\includegraphics[width=0.130\linewidth]{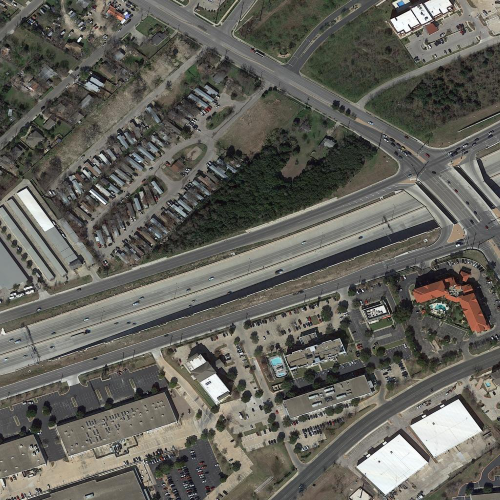}
		\label{fig:2}}    \hspace{-2.5mm}  
	\subfloat{
		\includegraphics[width=0.130\linewidth]{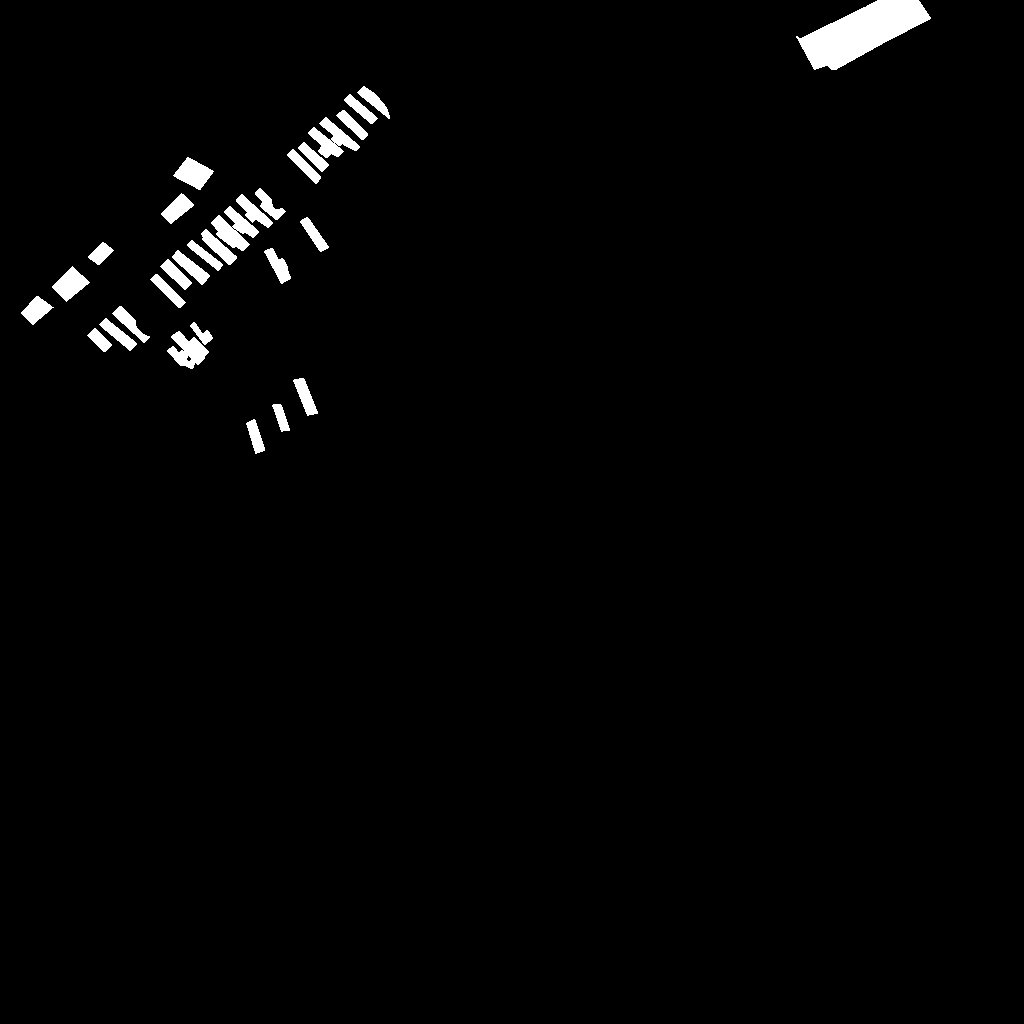}
		\label{fig:1} } \hspace{-2.5mm}  
	\subfloat{
		\includegraphics[width=0.130\linewidth]{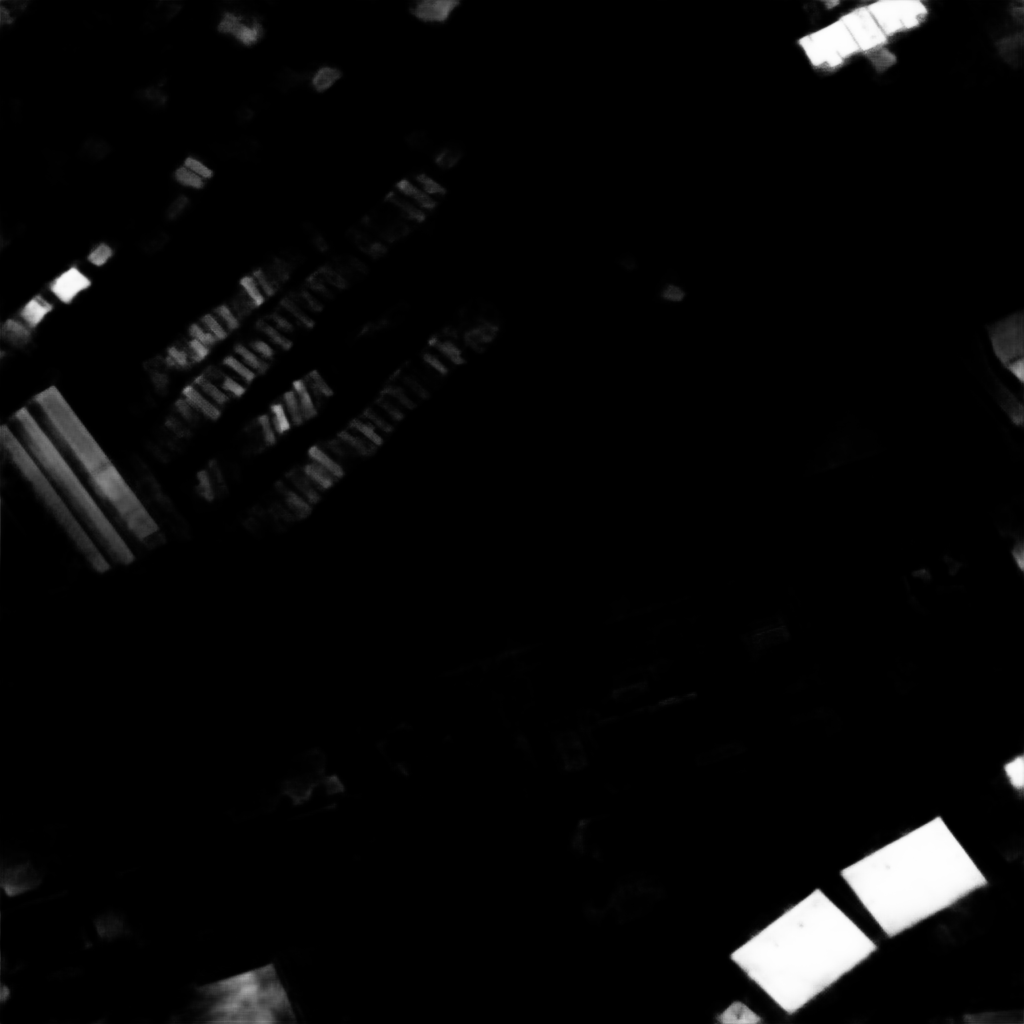}
	     \label{fig:2}}  \hspace{-2.5mm}       
	 \subfloat{
		\includegraphics[width=0.130\linewidth]{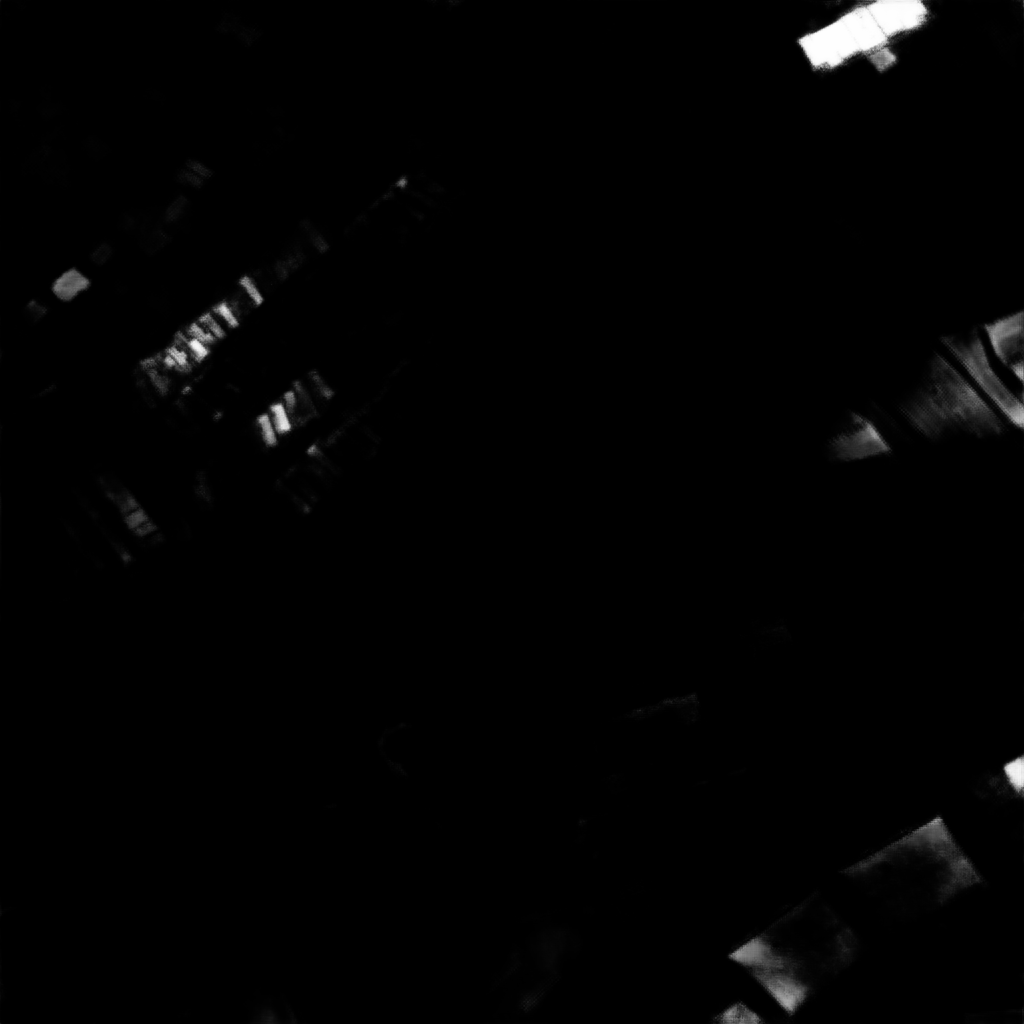}
	     \label{fig:2}}    \hspace{-2.5mm}  
	 \subfloat{
		\includegraphics[width=0.130\linewidth]{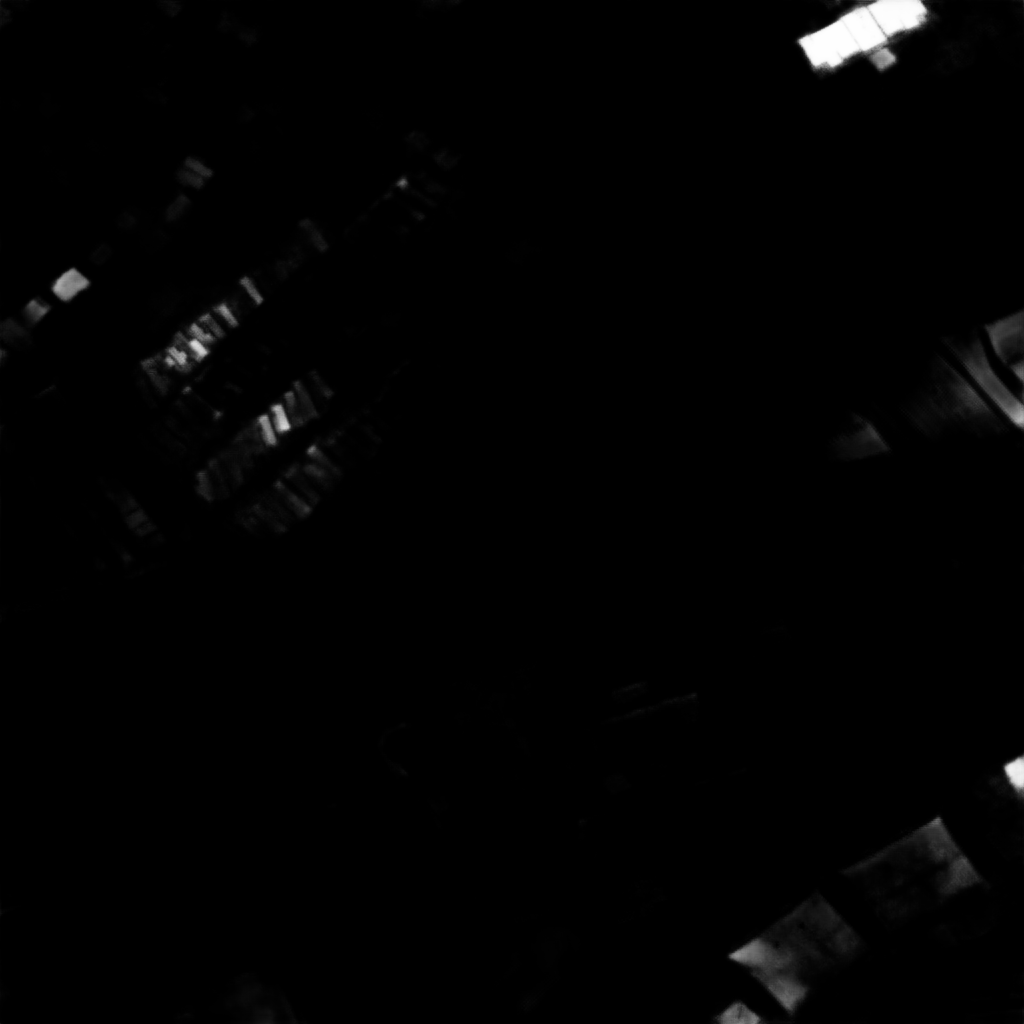}
	     \label{fig:2}}   \hspace{-2.5mm}   \\
	    	  \subfloat{
		\includegraphics[width=0.130\linewidth]{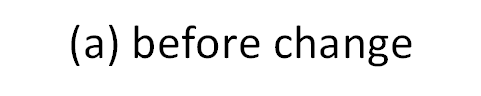}
		\label{fig:1} } \hspace{-2.5mm}  
	\subfloat{
		\includegraphics[width=0.130\linewidth]{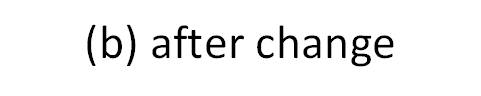}
		\label{fig:2}}    \hspace{-2.5mm}  
	\subfloat{
		\includegraphics[width=0.130\linewidth]{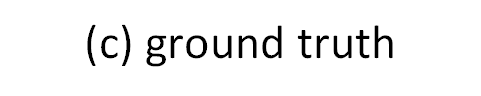}
		\label{fig:1} } \hspace{-2.5mm}  
	\subfloat{
		\includegraphics[width=0.130\linewidth]{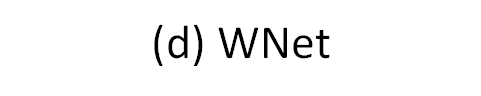}
	     \label{fig:2}}  \hspace{-2.5mm}       
	 \subfloat{
		\includegraphics[width=0.130\linewidth]{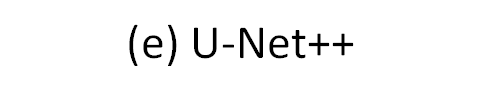}
	     \label{fig:2}}    \hspace{-2.5mm}  
	 \subfloat{
		\includegraphics[width=0.130\linewidth]{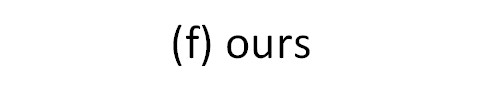}
	     \label{fig:2}}   \hspace{-2.5mm}   \\
	\caption{Building change detection results of different methods on LEVIR-CD dataset. (a) before change, (b) after change, (c) ground truth, (d) WNet, (e) U-Net++  and (f) our proposed BDANet. }
	\label{figImgCD}
\end{figure*}

2) \textit{Analysis of CDA}. To explore the correlations between pre- and post-disaster features, three CDA modules are used in the decoder of the network, which are represented as dconv1, dconv2 and dconv3 in Fig. \ref{figFramework}. This experiment investigates the influence of the number of CDA modules as well as the embedding position of CDA in the network.  Table \ref{tabAbCDA} shows that only slight performance gain is achieved when CDA is added in the convolutional layer of a small scale (dconv1). In contrast, a better improvement is obtained when CDA is added in the layer with a larger scale (dconv3). This is reasonable because more detailed information can be extracted when the feature maps are in a larger scale. When all three CDA modules are used in the network, around 0.6\% improvement is achieved compared with the vanilla network.

As discussed in Section \ref{SecMethod}-C, the CDA module is developed from the SE module \cite{royRecalibratingFullyConvolutional2019}. Therefore, in this experiment, we further evaluate the effectiveness of CDA when compared with the SE module. In the experiment, CDA is replaced with the channel (cha) and spatial (spa) attention modules of SE. As reported in Table \ref{tabAbSE}, only 0.1\% improvement is achieved when SE channel attention module is applied in the network. With both channel and spatial attention of SE, more gains are obtained in $F1_s$. Our proposed CDA provides a better result in comparison with all SE modules, which indicates that exploring the features between pre- and post-disaster images is effective in improving the damage assessment performance.

3) \textit{Analysis of CutMix.}
For the xBD dataset, \textit{minor damage} and \textit{major damage} are the most difficult classes based on the results in Table \ref{tabBase}. Therefore, in our proposed method, CutMix data augmentation is mainly focused on these two damage levels to generate more training samples for them. We further conduct experiments to investigate the influence of CutMix on different damage levels. Table \ref{tabAbCutmix} presents the overall scores when CutMix is applied on different damage levels. It is obvious that there is barely any improvement when CutMix is used on the entire dataset (i.e., considering all classes). In contrast, a significant improvement is achieved when only \textit{minor} and \textit{major} damage levels are considered in the data augmentation. This class-specific data augmentation strategy facilitates the network to learn better representations for those classes. 

\subsection{Computational Cost and Training Efficiency}
We also compare the parameter size, computational cost (FLOP) and inference speed of the proposed overall framework, the vanilla network with MFF and CDA modules. The CutMix strategy is used in data pre-processing stage and thus it does not consume additional computational resources in the network.   
As shown in Table \ref{tabParam}, the MFF module adds a little more computational cost since more convolutional operations are applied. Although the CDA module has a complicated architecture, it uses few convolutional operations. Therefore, it includes less than 1M additional parameters.  This results in only a slight increase in computational cost (107.6 vs 92.9 GFLOPs). Moreover, our proposed framework only increases less than 2M parameters and about 60G FLOPs as compared with the vanilla network.  We also calculate the average inference time (s) for each image ($1024\times1024$ pixels) in the testing phase. The additional runtime (0.016s) is negligible. 

In addition, we evaluate the training efficiency of using weights from Stage 1. As discussed in Section \ref{SecMethod}, the network weights from Stage 1 are used as initial weights in the network of Stage 2 for fine tuning. Fig. \ref{figCurve} plots the validation accuracy and loss versus the number of epochs. It is obvious that a higher validation accuracy and a smaller loss value are achieved at the first epoch when adopting the weights from Stage 1 as initialization. With the number of epochs increasing, the curve of using weights from Stage 1 converges faster. Fig. \ref{figCurve} shows that at least 5 epochs are saved when using the weights from Stage 1.


\subsection{Evaluation on Building Change Detection Dataset}
\textcolor{black}{As more datasets on building damage assessment are difficltut to acquire, a building change detection dataset LEVIR-CD \cite{chenSpatialTemporalAttentionBasedMethod2020} is used for evaluation of our proposed framework. These images are captured in several cities of Texas, US from 2002 to 2018. The dataset collected 637 high resolution images (0.5m per pixel) with a size of $1024\times 1024$ via Google Earth. Among these images, 509 images pairs are used for training and 128 image pairs are used for testing.}

\textcolor{black}{Some change detection-based methods, including WNet \cite{houWNetCDGANBitemporal2020} and U-Net++ \cite{pengEndtoEndChangeDetection2019}, are uesd in comparison with BDANet. For the change detection task, only the network in Stage 2 is used for training. The results of different methods on LEVIR-CD are reported in Table \ref{tabCD}. From the Table, we can observe that the proposed method achieved best score on three indexes (precision, recall and F1), indicating that BDANet is also effective on other dataset. Fig. \ref{figImgCD} provides some visual results of different change detection methods. We can also observe that the proposed BDANet provides better performance compared with other methods.}

\section{Conclusion}
\label{SecConclusion}
In this paper, a two-stage CNN-based learning framework, named BDANet, is proposed for building damage assessment from satellite images. In Stage 1, a U-Net architecture is used for building segmentation, and the output is used for guiding the building locations in damage assessment. In Stage 2, the MFF and CDA modules are used for further improving the assessment performance. MFF is used to learn features from various scales of images, which enables the network to build a robust feature representation. The CDA module is used to explore the correlations between pre- and post-disaster features and enhance the feature representation capability. Moreover, a data augmentation strategy CutMix is employed to mitigate the challenge of difficult classes. Experimental results show that significant improvements can be achieved when CutMix is applied on difficult damage levels. The proposed method also yields state-of-the-art building damage assessment performance on the xBD dataset.



%

\ifCLASSOPTIONcaptionsoff
  \newpage
\fi



\bibliographystyle{IEEEtran}
\bibliography{bib/IEEEabrv,bib/cite,bib/damageAssess}
\end{document}